\documentclass[journal]{IEEEtran} 

\usepackage{cite}
\usepackage{amsmath,amssymb,amsfonts}
\usepackage{graphicx}
\usepackage{xcolor}
\usepackage{multirow}

\usepackage{graphicx}
\usepackage{graphics} 
\usepackage{mathptmx}      
\usepackage{latexsym}
\usepackage{algorithm}
\usepackage{algorithmicx}
\usepackage{algpseudocode}
\usepackage{subcaption}
\usepackage{amsmath}
\usepackage{newtxtext,newtxmath}
\usepackage{hyperref}
\usepackage{cleveref}

\usepackage{epsfig} 
\usepackage{mathptmx} 
\usepackage{times} 
\usepackage{caption}
\usepackage{booktabs}
\usepackage{colortbl}
\usepackage{wrapfig}
\colorlet{lightgray}{gray!20}
\newcommand{\rx}[1]{\textcolor{black}{#1}}
 \renewcommand{\paragraph}[1]{
    \vspace{2mm}
     \noindent\textbf{#1} 
 }
\begin{document}
\title{The OpenCDA Open-source Ecosystem for Cooperative Driving Automation Research}

\author{Runsheng Xu, Hao Xiang, Xu Han, Xin Xia, Zonglin Meng, 
 Chia-Ju Chen
        and Jiaqi Ma*,~\IEEEmembership{Member,~IEEE}
\thanks{Runsheng Xu, Hao Xiang, Xu Han, Xin Xia, Zonglin Meng, 
 Chia-Ju Chen and Jiaqi Ma are with the Department
of  Civil and Environmental Engineering, University of California, Los Angeles, CA, 90024, USA}
\thanks{* corresponding author: jiaqima@ucla.edu.}

}

\markboth{Journal of \LaTeX\ Class Files}%
{Shell \MakeLowercase{\textit{et al.}}: Bare Demo of IEEEtran.cls for IEEE Journals}

\maketitle

\begin{abstract}
Advances in Single-vehicle intelligence of automated driving has encountered great challenges because of limited capabilities in perception and interaction with complex traffic environments. Cooperative Driving Automation~(CDA) has been considered a pivotal solution to next-generation automated driving and smart transportation. Though CDA has attracted much attention from both academia and industry, exploration of its potential is still in its infancy.
In industry, companies tend to build their in-house data collection pipeline and research tools to tailor their needs and protect intellectual properties. Reinventing the wheels, however, wastes resources and limits the generalizability of the developed approaches since no standardized benchmarks exist.
On the other hand, in academia, due to the absence of real-world traffic data and computation resources, researchers often investigate CDA topics in simplified and mostly simulated environments, restricting the possibility of scaling the research outputs to real-world scenarios. 
Therefore, there is an urgent need to establish an open-source ecosystem~(OSE) to address the demands of different communities for CDA research, particularly in the early exploratory research stages, and provide the bridge to ensure an integrated development and testing pipeline that diverse communities can share. 
In this paper, we introduce the OpenCDA research ecosystem, a unified OSE integrated with a model zoo, a suite of driving simulators at various resolutions, large-scale real-world and simulated  datasets, complete development toolkits for benchmark training/testing, and a scenario database/generator. We also demonstrate the effectiveness of OpenCDA OSE through example use cases, including cooperative 3D LiDAR detection, cooperative merge, cooperative camera-based map prediction, and adversarial scenario generation.

\end{abstract}


\begin{IEEEkeywords}
cooperative driving automation, automated driving, intelligent transportation system, digital twin
\end{IEEEkeywords}

\IEEEpeerreviewmaketitle
\section{Introduction}
\label{sec:1} 


The advent of automated driving technology has garnered substantial attention and investment in recent years, with many experts positing that it has the potential to revolutionize the transportation industry~\cite{xu2021opencda,valiente2022robustness,hua2019hierarchical,chen2019comprehensive,calvert2020conceptual,shet2021cooperative,ferrara2022multi,liu2021automated,wang2017parallel}. Automated vehicles possess the capability to significantly enhance road safety, alleviate traffic congestion, and increase mobility for individuals who are unable to operate a vehicle, such as the elderly or disabled. Furthermore, the implementation of automated vehicles has the potential to reduce the number of cars on the road, thus leading to decreased pollution and energy consumption.

Despite the significant progress that has been made in the field of automated driving, single-vehicle perception systems are still prone to occlusion and are limited in their ability to detect distant objects~\cite{wang2020v2vnet,xu2022opv2v,bai2022survey,hu2022investigating,li_message,blumenkamp2022learning,ma2021analysis}.  Furthermore, the integration of automated vehicles into a mixed-autonomy transportation system, where they must interact with human-driven vehicles, pedestrians, complicated traffic rules, and various types of infrastructure, poses significant difficulties in making safe decisions~\cite{peng2021learning,hu2020seasondepth,ma2021analysis}. These challenges remain a major obstacle to the large-scale deployment of automated driving technology in real-world scenarios.


Cooperative Driving Automation (CDA) has recently gained significant attention as a solution to the challenges faced by individual automated vehicles, such as occlusion and limited perception range~\cite{9497693,9298806,9415170,8671766,9857660,9762043,9068410,9684293,9851671,9495943}. According to SAEJ3216~\cite{SAE}, CDA refers to the use of machine-to-machine communication to enable cooperation among multiple entities, such as vehicles, pedestrians, and infrastructure components, equipped with capable communication technologies. With the support of CDA, automated vehicles can see through occlusion, extend their perception range~\cite{xu2022v2xvit,hu2022where2comm}, and make more intelligent decisions~\cite{viana2021comparison,xu2021opencda}. One of the early programs to research CDA is the CDA/CARMA program hosted by the U.S. Department of Transportation, which initiated SAEJ3216~\cite{lochrane2020carma}.

CDA has the potential to bring significant benefits to the field of automated driving. However, CDA is still in its early stages of development. In academia, researchers in university labs often have limited access to real-world traffic data and computational resources, so they typically study CDA in simplified simulation environments. This makes it difficult to apply their research findings to real-world scenarios. Additionally, most research studies utilize simulation capabilities that are not comparable between studies, hindering the ability to compare and evaluate the performance of different approaches. On the other hand, while companies in the industry typically have access to rich resources, they tend to primarily focus on current single-vehicle technology rather than investing in CDA research. Furthermore, those that do invest in CDA often keep their data and pipelines confidential. To promote the healthy and rapid development of CDA, the creation of an open-source ecosystem (OSE) is essential. Such an OSE should be able to: 1) meet the requirements of various stakeholders; 2) facilitate the integration of development and testing pipelines across different communities; 3) furnish a comprehensive toolset that can be utilized by various communities for the development of functionalities and seamless integration with other pertinent modules throughout the automated vehicle-transportation interaction process.

To this end, we introduce the OpenCDA Ecosystem, which is grown from our previous simulation tool OpenCDA~\cite{xu2021opencda}. The OpenCDA ecosystem is a comprehensive OSE that includes a collection of pre-trained models, a range of simulators for driving and traffic at different levels of detail, benchmark datasets for training and testing, and a scenario database and generator for CDA research. The OpenCDA ecosystem comprises four main components:  \textbf{toolkits}, \textbf{scenario databases}, \textbf{data}, and \textbf{communities}:
\begin{itemize}
    \item \textbf{Toolkits}: The OpenCDA ecosystem offers a comprehensive set of toolchains to assist communities in prototyping and developing their CDA systems and algorithms. It includes a simulation tool, a deep learning framework, and libraries of models that include state-of-the-art algorithms for various automated driving tasks.
    \item \textbf{Data}: The OpenCDA ecosystem contains both simulated and real-world large-scale datasets to support offline training for cooperative perception algorithms.
    \item \textbf{Scenarios}: OpenCDA offers a scenario database, generation, and man-
    management tools, including various rule-based standard scenarios
    for testing different cooperative driving applications as benchmarks.  
    \item \textbf{Community}: OpenCDA has a comprehensive and well-structured plan to establish and nurture a thriving community of researchers and engineers with diverse backgrounds and skill sets to facilitate the development of CDA.
\end{itemize}

The rest of the paper is organized as follows: we will first introduce the literature in the CDA field in \cref{sec:2}, then we will describe the details of each component of our OpenCDA ecosystem  in \cref{sec:3}. Afterward, we will conduct several case studies to demonstrate how the OpenCDA ecosystem facilitates different research needs. Eventually, we will conclude the work and depict the future orientations.


\section{Related Work}
\label{sec:2} 
\begin{figure*}[!t]
\centering
\footnotesize
\includegraphics[width=0.80\textwidth]{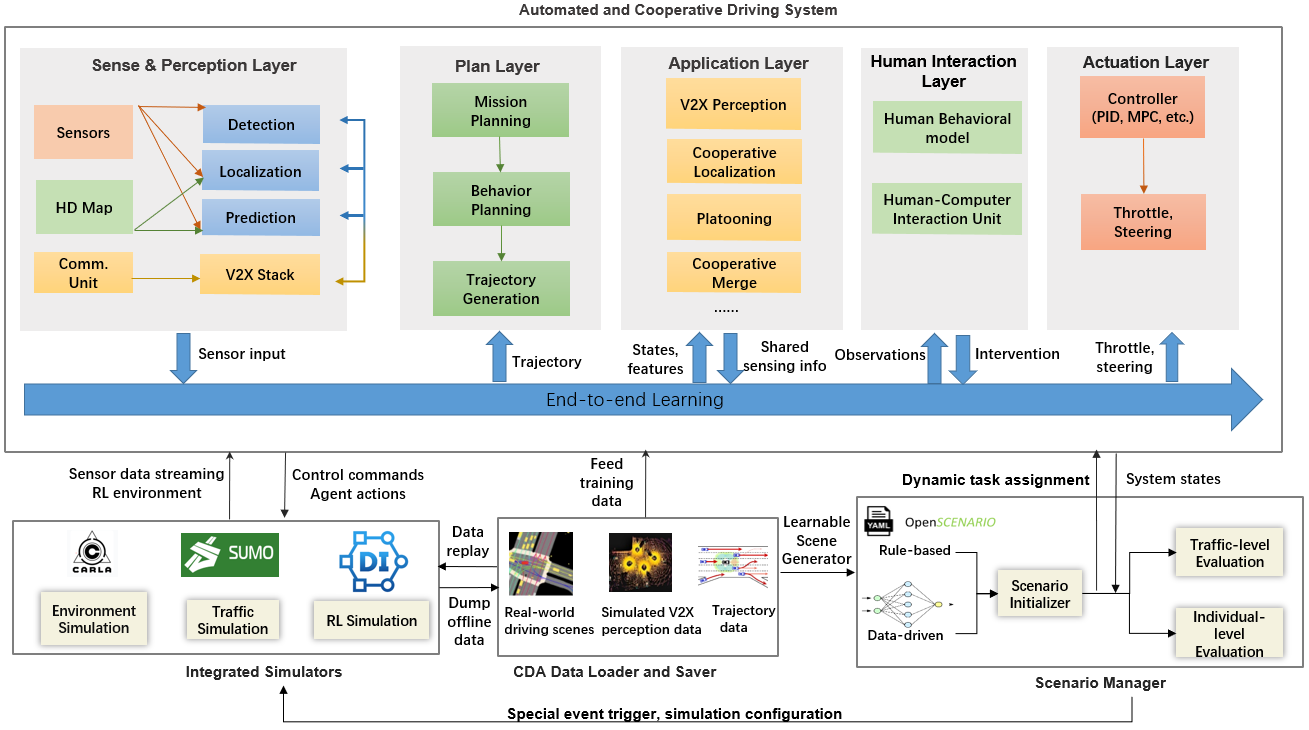}
\caption{Overview of the OpenCDA simulation tool architecture}
\label{fig:opencda}
\end{figure*}

\subsection{Simulators}
Conducting field experiments in the field of automated driving can present significant challenges, particularly in terms of cost and safety. To mitigate these issues, researchers and engineers often first prototype their systems in simulated environments before implementing them in the real world. Among the available simulators, CARLA~\cite{dosovitskiy2017carla} has been a popular choice, offering a free, open-source platform that aims to accelerate the development of new automated driving technologies. Furthermore, other simulators like SUMO~\cite{SUMO2018} and Flow~\cite{wu2017flow} are also widely used, but they are typically focused on traffic-level simulation and do not support regular automated driving components and cooperative driving automation. Our previous work, OpenCDA~\cite{xu2021opencda}, aims to fill this gap by providing an integrated framework that supports full-stack CDA software development in simulation and can be used to simulate various scenarios with the support of various communities.

\subsection{Cooperative Driving Automation}
\rx{Cooperative driving automation allows vehicles to communicate with each other and with the infrastructure around them to improve safety and increase efficiency on the road. The technology uses a combination of sensors, including radar, lidar, and cameras, as well as wireless communication systems, such as dedicated short-range communication (DSRC) and cellular networks, to enable vehicles to share information about their intents and observation.} Several CDA applications have emerged in the past decade and hold great premise in impacting the transportation systems, such as Cooperative Adaptive Cruise Control~(CACC), platooning, cooperative merge, cooperative perception, eco-drive at signalized intersections. The applications also evolve quickly as supporting technologies, such as computer vision and vehicular communication, advance. 

For example, CACC, i.e., single-lane decentralized ad-hoc operations of multiple vehicles that closely follow each other, has been widely investigated~\cite{Liu2018,Shladover2015COOPERATIVEAC,9352029}. Although CACC can improve the safety and comfort of driving, it is only applied longitudinally, and each vehicle can only communicate with the front vehicle. In contrast, another important CDA application, i.e., platooning, enabled by higher levels of vehicle automation, can handle multi-lane scenarios and allow the members to cooperate with both the front vehicle and the leader~\cite{han2022strategic}. To implement such a system, each platoon member communicates simultaneously via an established principle that shares intention and operating status that serve as the input of longitudinal (i.e., speed regularization) and lateral control (i.e., lane change). The platooning algorithms can either be distributed~\cite{zhang2020distributed,zhou2022event} or centralized~\cite{guo2021scopto,sokolov2017maximization}.  

Recently, another example application that has receive much attention is cooperative perception, which builds upon recent progress in computer vision and ADS perception. Cooperative perception studies how to efficiently fuse visual cues from neighboring vehicles or infrastructure~\cite{xu2022v2xvit,cai2022analyzing,li2021learning,xu2022bridging,chen2022model,yuan2022keypoints,bai2022vinet,li2022multi}. Based on the vision information sharing strategy, it can
be divided into three categories: 1) early fusion, where raw data is shared and gathered to form a holistic view, 2) intermediate fusion, where intermediate neural features are extracted based on each agent’s observation and then transmitted; and 3) late fusion, where detection outputs (e.g., 3D bounding box position, confidence score) are circulated. By utilizing OpenCDA~\cite{xu2021opencda} simulation tool, OPV2V~\cite{xu2022opv2v} collects the first large-scale Vehicle-to-Vehicle~(V2V) simulated perception dataset and proposes the first cooperative detection framework OpenCOOD~\cite{opencood_repo}. Since then, an increasing number of related works that use OpenCOOD as the base have shown up and brought significant impacts to the CDA field, including V2X-Vit~\cite{xu2022v2xvit}, CoBEVT~\cite{xu2022cobevt}, Where2Comm~\cite{hu2022where2comm}, Map Container~\cite{jiang2022map}, CoAlign~\cite{lu2022robust}, and AdaFusion~\cite{qiao2022adaptive}. 

The Federal Highway Administration's (FHWA) CARMA Program~\cite{lochrane2020carma} is a leading research initiative that utilizes emerging technologies in automation and cooperation to improve transportation systems management and operations. The program has developed software platforms for both vehicles and infrastructure, as well as tools for full-scale vehicle software simulation and testing. In partnership with the CARMA Program, OpenCDA - an open-source ecosystem - makes a unique contribution by providing early-stage development and testing opportunities for users to evaluate their customized algorithms in terms of specific tasks, such as object detection accuracy and overall pipeline assessment for traffic safety.


\section{The OpenCDA Ecosystem}
\label{sec:3} 

In this section, we will present the details of the major components of the OpenCDA ecosystem (as of the publication date of this paper, i.e., January 2023) and then explain how they interact with each other through the interfaces to form an ecosystem.

\subsection{Development Toolkits}
The development toolkits are a set of software routines and utilities that assists the communities in prototyping and developing their CDA systems and algorithms.The toolkits contain a simulation tool, a deep learning framework OpenCOOD~\cite{opencood_repo}, and a model zoo that contains state-of-the-art~(SOTA) algorithms of different automated driving tasks.

\subsubsection{OpenCDA Simulation Tool}
OpenCDA\footnote{code link: https://github.com/ucla-mobility/OpenCDA} started as an open co-simulation-based simulation tool  integrated with prototype cooperative driving automation pipelines as well as
regular automated driving components (e.g., perception, localization, planning, control). This simulation tool is the core element of the whole ecosystem, as it 
enables CDA evaluation in a CARLA + SUMO co-simulation environment and provides a rich library of source codes of CDA research pipelines. In collaboration with U.S.DOT CDA Research and the FHWA CARMA Program~\cite{lochrane2020carma}, OpenCDA, as an open-source project, is designed and built to support
early-stage fundamental research for CDA research and development.

As figure~\ref{fig:opencda} depicts, the simulation tool has four major parts: simulator integration, cooperative driving system, scenario manager, and CDA data loader and saver:

\noindent\textbf{Integrated simulation}: OpenCDA integrate CARLA~\cite{dosovitskiy2017carla} for realistic environment rendering, SUMO~\cite{SUMO2018} for realistic background traffic, and DI-Engine~\cite{ding} to provide reinforcement learning environment. These simulators together will supply the sensor data streaming and necessary environment information to the cooperative driving system.

\noindent\textbf{Automated and Cooperative Driving System}: OpenCDA builds on the philosophy that CDA functionalities need to build on and also enhance automated driving technologies of single vehicle intelligence. Therefore, OpenCDA prototypes the entire regular automated driving systems software and adds modular CDA components (e.g., V2X communication, cooperative perception, platooning) to the driving automation system. For example, the percpetion backbones of 3D LiDAR or camera perception algorithms starts within single vehicle automation and cooperative perception builds on top of it by exploring different strategies of information sharing and fusion (e.g., intermediate or late fusion). Platooning modules still relies on signle vehicle plan and control modules but only interfaces with platooning strategic and tactical decision making modules to output desired trajectories and maneuvers that not only guarantee single vehicle safety but also satisfies platooning requirements.

Specifically, OpenCDA purely utilizes Python to prototype the cooperative driving system, which consists of the sensing layer, plan layer, application layer, and actuation layer. After receiving the sensor information from simulators, the sensing layer will perform object detection, road topology prediction, and localization. Then the sensing and localization information will be delivered to the planning layer for behavior plan and trajectory generation. The outputs of the planning module will be fed into the actuation layer, where the control commands are finally generated and sent back to the simulator to update. When multiple AVs meet, the sensing layer's communication unit will start working and activating the application layer, where cooperative functionalities such as cooperative perception and platooning will replace the default perception and planning modules. Besides the classic modular pipeline, OpenCDA also supports end-to-end driving, i.e., the control commands are directly generated from the raw sensor measurements without any intermediate outputs. 

\rx{The cooperative driving system component also includes a special layer called the human interaction layer, which is made up of two sub-units: a human behavioral model and a human-computer interaction unit. This layer receives inputs from all other layers of the cooperative driving system, as well as human behavior inputs in response to generated driving scenarios. The interaction mechanism unit then outputs commands to the plan and actuation layers. The human operator model is designed to operate in two modes: simulation-based and experimental-based. In simulation-based mode, model- and simulation-based Human Reliability Analysis (HRA) methods, such as the Information-Decision-Action Crew (IDAC) operator model~\cite{chang2007cognitive}, are used to simulate human performance. In the experimental-based mode, humans receive inputs from the automated vehicle system. The experimental setup adheres to human factors and ergonomic principles with a level of fidelity suitable for specific studies. The interaction model allows for human intervention in AV operation at two levels: remote commands (e.g., waypoints, remote stops) and local vehicle control (e.g., take-over requests, disengagement mechanisms). This flexibility allows for the simulation of different operational scenarios and the assessment of task-sharing strategies between the human operator and the AV system under various operating conditions. The setup can also be used to collect data on human factors and behavior performance influencing factors in time-constrained monitoring and intervention scenarios.}

\noindent\textbf{Scenario Manager}: The scenario manager mainly has three roles: 1) It depicts the scenarios at the beginning of the simulation, including weather, traffic states, testing locations, etc. 2) It assigns the dynamic driving tasks to different AVs, including start positions, destinations, and the category of the cooperating maneuvers need to be operated. 3) It provides both traffic level and individual level evaluation after each run.

\noindent\textbf{CDA Data Loader and Saver}: In OpenCDA, we collect the simulated perception and traffic data in a similar approach as the real world. The connected AVs will drive simultaneously in the same scenario, and their multi-modal sensors will keep capturing the surrounding information. The CDA data saver will provide convenient APIs to convert the online sensor data to an appropriate format and save it to the disk, which can be directly used for deep learning training. \rx{Please refer to the Section \ref{sec:data_management} to see more details of the data saved by the CDA Data Saver.}Moreover, the CDA data loader will provide interfaces to replay the collected simulated data to evaluate how the trained model performs when integrating with the full-stack system. It is also worth noting that the CDA data loader can also replay real-world automated driving data if it has high-definition maps~(HDMap) in osm format with high-precision GPS data. 

\noindent\textbf{Component Interactions}: \rx{The components of the OpenCDA simulation tool interact with each other in an organized manner.  The integrated simulators provide sensor data streaming, set up the reinforcement learning environment, and control the background traffic for each CAV. Each CAV is controlled by its own cooperative driving system, which receives sensor data and environment metadata and generates driving decisions and control commands based on this information. These control commands are then sent back to the simulators to update the simulation environment. The scenario manager component defines the dynamic driving task and scenario for each CAV, and upon completion of the simulation, it collects information from each cooperative driving system to provide driving evaluations. If the CDA data loader component is activated, the scenarios can be directly loaded from offline simulated or real-world data, bypassing the scenario manager. During the simulation, each CAV can also use the CDA data saver component to dump collected sensor data and create a new simulation dataset.}

\noindent\textbf{Development Status}: \rx{The OpenCDA simulation tool is continually being updated and expanded with new features. Currently, several components, such as the Human Interaction Layer, the End-to-end Learning pipeline, and the reinforcement learning environment integration in Fig.~\ref{fig:opencda}, are still under development. Research teams from various universities across different countries are actively working to improve these components. For example, the team from UC Davis is integrating a multi-agent reinforcement learning framework into OpenCDA, while scholars from Nankai University are using OpenCDA in distributed systems research. Additionally, the team from the Technical University of Munich is working to make OpenCDA compatible with the NS3 network simulator.}

\begin{figure*}[!t]
\centering
\footnotesize
\includegraphics[width=0.60\textwidth]{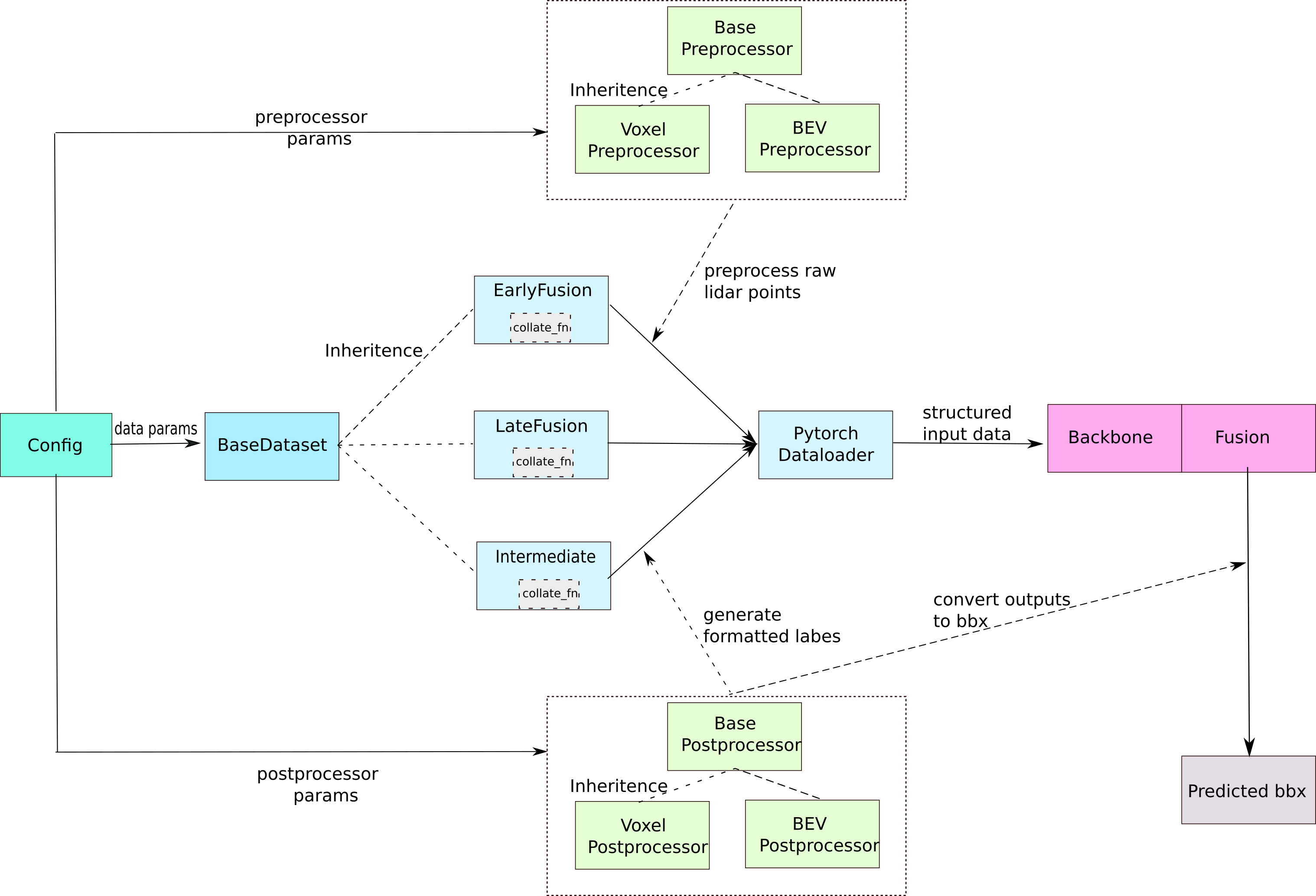}
\caption{OpenCOOD pipeline for cooperative perception.}
\label{fig:opencood}
\end{figure*}

\subsubsection{OpenCOOD}
OpenCOOD~\footnote{https://github.com/DerrickXuNu/OpenCOOD} is the first code framework using PyTorch~\cite{NEURIPS2019_9015} for cooperative detection and segmentation, foundation for cooperative perception research. Unlike popular 3D detection frameworks such as OpenPCDet~\cite{openpcdet2020} and MMDetection3D~\cite{mmdet3d2020}, OpenCOOD deals with multi-agent perception data, where multi-view sensor data streaming will be provided for the same scene simultaneously. As figure~\ref{fig:opencood} reveals, there are 6 major components:
\begin{itemize}
    \item \textbf{Config system}: OpenCOOD incorporates modular and inheritance design into the config system to enable users conveniently modify the model/training/inference parameters. Specifically, we use YAML files to configure all the important parameters. 
    \item \textbf{Dataset class}: Opencood has 3 different classes that inherit from the base dataset class to convert the multi-agent data into formats that common fusion strategies~(i.e., early fusion, late fusion, and intermediate fusion) can use.
    \item \textbf{Preprocessor and backbone}: The preprocessor and backbone together  will convert the raw point clouds to voxels or 2D bird-eye-view features. Please refer to \cref{subsec:model} to see the backbones this framework supports.
    \item \textbf{Multi-agent fusion}: The fusion module will aggregate the shared raw point cloud, bounding boxes, or neural features from different vehicles to output a holistic fused feature. Refer to \cref{subsec:model} to check the state-of-the-art fusion models we support.
    \item \textbf{Postprocessor}: The postprocessor will convert the model's outputs to the correct bounding box formats that can be utilized for visualization and evaluation.
\end{itemize}
\rx{Since its initial release, OpenCOOD has gained a significant amount of support from various research teams who have contributed to the model zoo and added support for additional datasets and modalities. For instance, the team from Shanghai Jiaotong University has added new models like Where2Comm~\cite{hu2022where2comm} and When2Comm~\cite{liu2020when2com} to OpenCOOD, while teams from Queen's University and Leibniz University Hannover have added support for datasets such as CODD~\cite{arnold2021fast} and multi-modality functionality to the platform.}

\subsubsection{Model Libraries}
\label{subsec:model}
The OpenCDA ecosystem offers a vast collection of models, including both traditional algorithms and deep learning models, which come pre-trained and can be utilized to facilate the development of CDA. This model library can be utilized not only within the OpenCDA ecosystem but also on other platforms. Upon this time, the model library includes the following:
\begin{itemize}
    \item\textbf{3D LiDAR-based Object Detection}: VoxelNet~\cite{zhou2018voxelnet}, PointPillar~\cite{lang2019pointpillars}, Pixor~\cite{yang2018pixor}, SECOND~\cite{yan2018second}.
    \item\textbf{Cooperative Perception}: V2X-ViT~\cite{xu2022v2xvit}, AttFuse~\cite{xu2022opv2v}, CoBEVT~\cite{xu2022cobevt}, Where2Comm~\cite{hu2022where2comm}, FPV-RCNN~\cite{yuan2022keypoints}, F-Cooper~\cite{chen2019f}, DiscoNet~\cite{li2021learning}, V2VNet~\cite{wang2020v2vnet}.
    \item\textbf{Localization}: Kalman Filter~\cite{chui2017kalman}, Extended Kalman Filter~\cite{einicke1999robust}.
    \item\textbf{Planning}: A* Searching, Spline Cubic Polynomial trajectory generation.
    \item\textbf{Platooning}: Finite state machine~\cite{han2022strategic}.
    \item\textbf{Control}: PID control~\cite{johnson2005pid}, MPC~\cite{afram2014theory}.
\end{itemize}

\begin{figure}[!t]
\centering
\subfloat[OPV2V]{%
  \includegraphics[width=0.9\columnwidth]{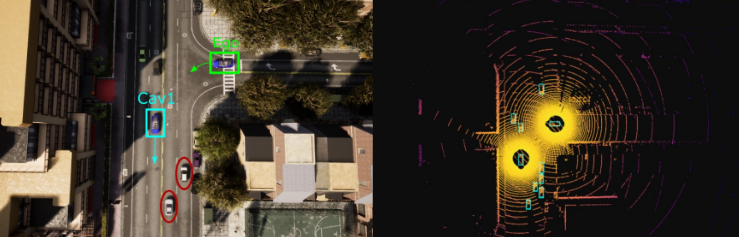}%
}
\hfil
\subfloat[V2XSet]{%
  \includegraphics[width=0.9\columnwidth]{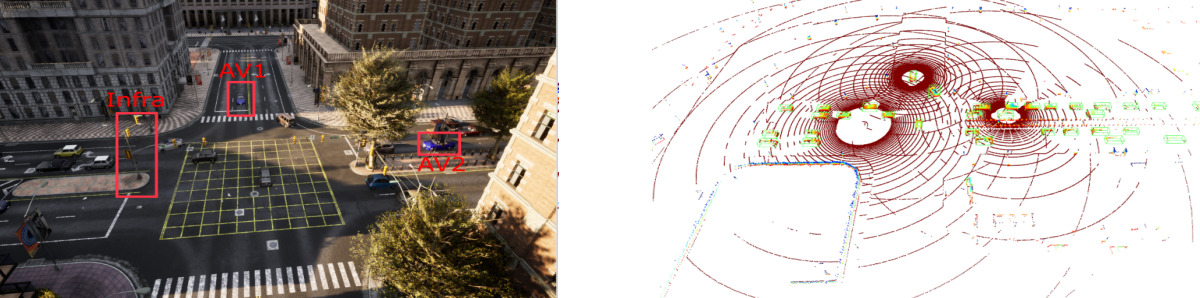}%
}
\hfil
\subfloat[Real-world data. \textit{Left Up}: The aggregated 3D LiDAR. \textit{Right Up}: The annotated HDMap, where the stars indicate the position of the two collection vehicles. \textit{Bottom Row}: The front cameras of the two vehicles.]{%
  \includegraphics[width=0.9\columnwidth]{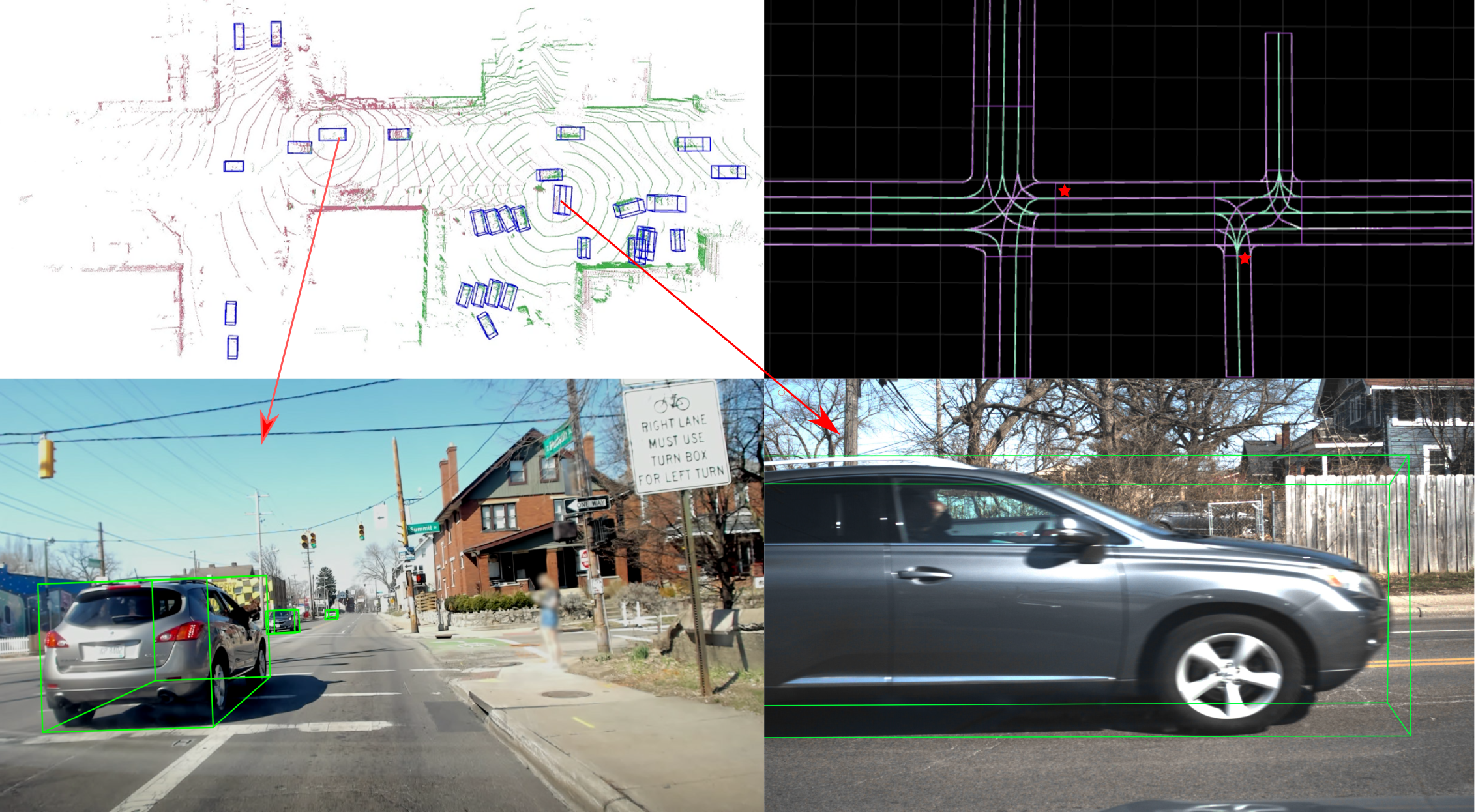}%
}
\caption{Data samples from different datasets in the OpenCDA ecosystem.}
\label{fig:data_sample}
\vspace{-4mm}
\end{figure}

\begin{figure}[!t]
\centering
\subfloat[Real-world scenario]{%
  \includegraphics[width=0.9\columnwidth]{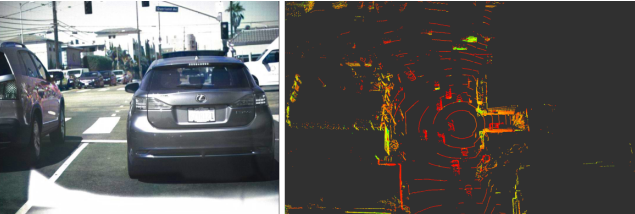}%
}
\hfil
\subfloat[Corresponding digital twin in CARLA]{%
  \includegraphics[width=0.9\columnwidth]{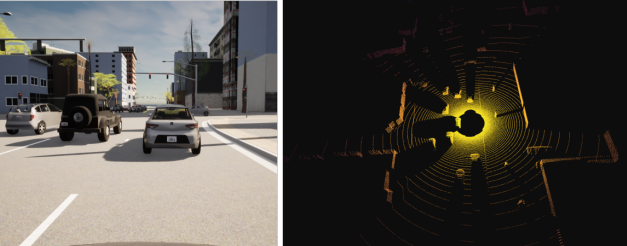}%
}

\caption{\rx{Our OpenCDA ecosystem can create a digital twin in CARLA for the real-world dataset for Culver City in Los Angeles.}}
\label{fig:digit_twin}
\vspace{-4mm}
\end{figure}

\begin{figure*}[!t]
\centering
    \begin{subfigure}[c]{0.24\linewidth}
        \centering{\includegraphics[width=1\linewidth]{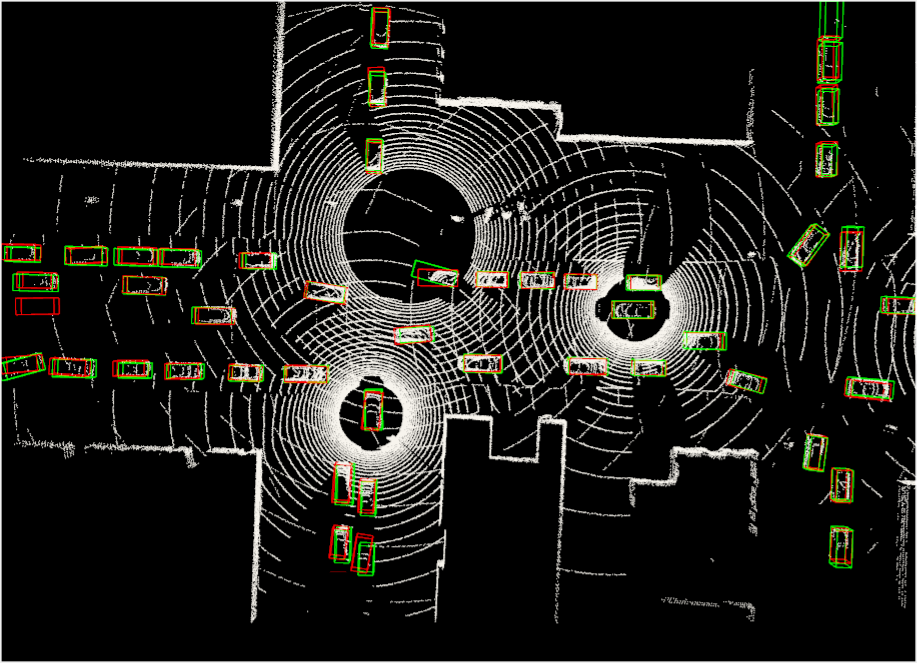}}
        \centering{\includegraphics[width=1\linewidth]{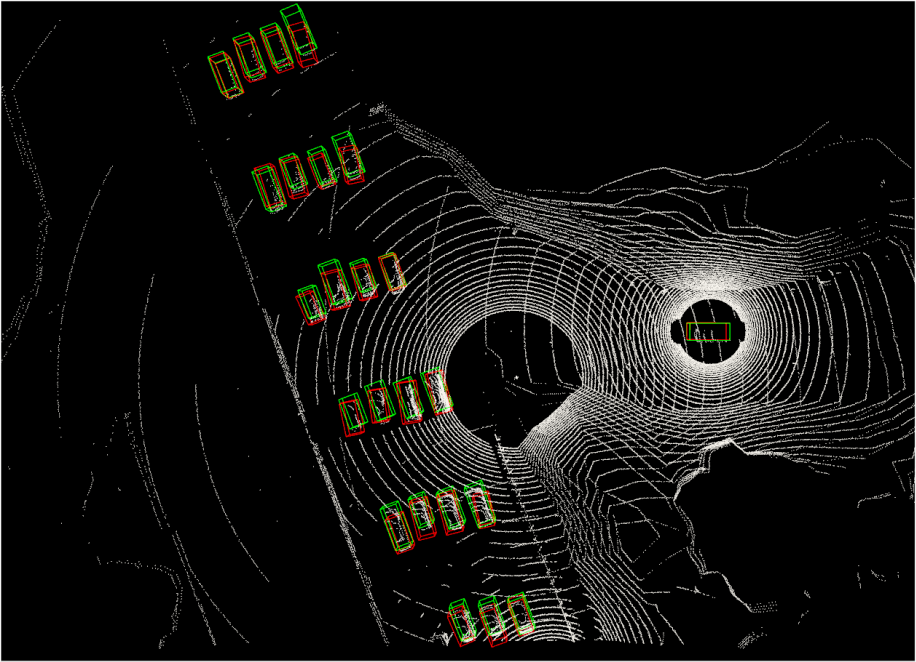}}
        \caption{OPV2V~\cite{xu2022opv2v}}
        \label{fig:output-a}
    \end{subfigure}
    \begin{subfigure}[c]{0.24\linewidth}
        \centering{\includegraphics[width=1\linewidth]{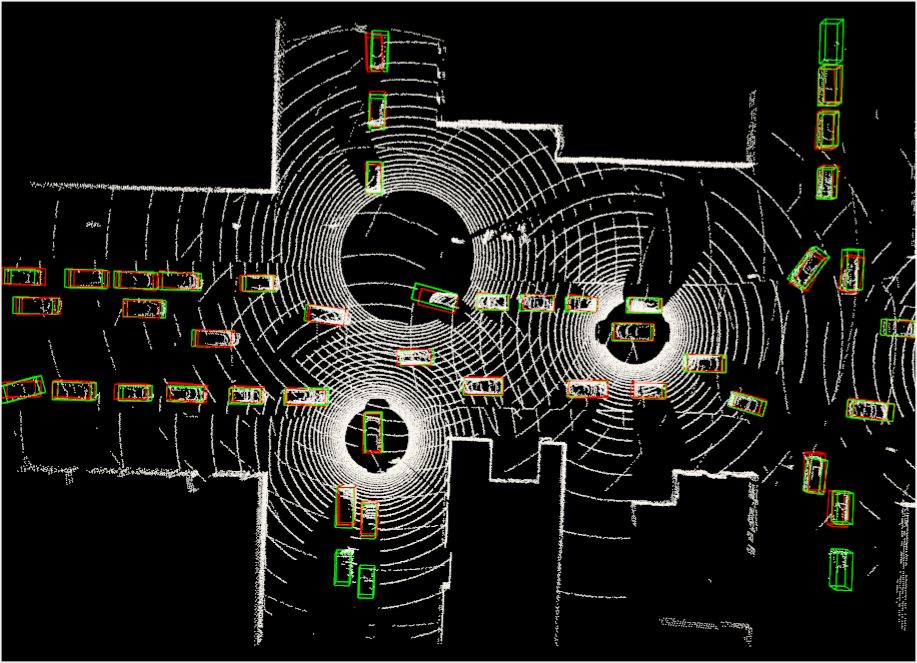}}
        \centering{\includegraphics[width=1\linewidth]{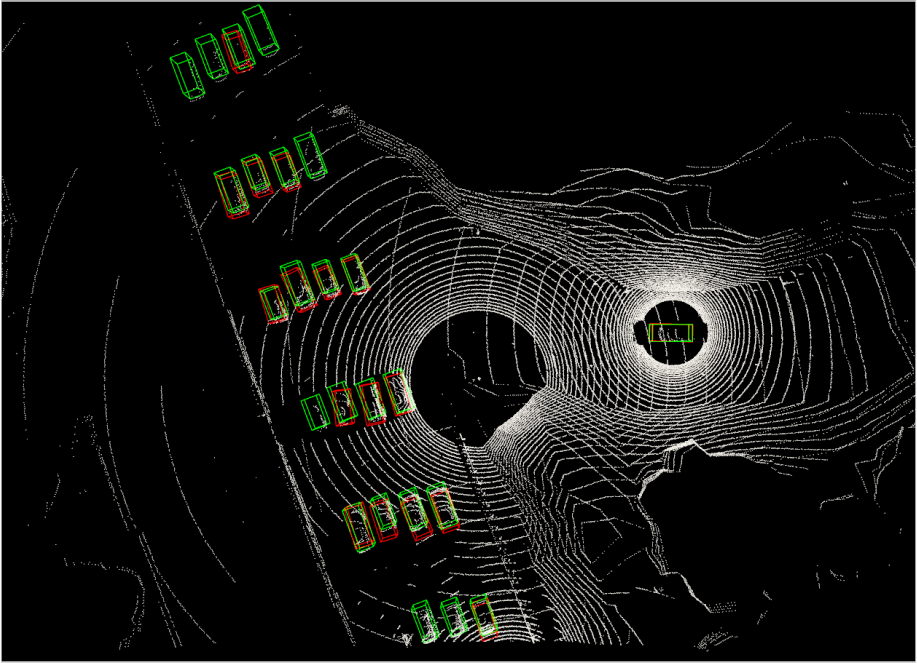}}
        \caption{V2VNet~\cite{wang2020v2vnet}}
        \label{fig:output-b}
    \end{subfigure}
    \begin{subfigure}[c]{0.24\linewidth}
        \centering{\includegraphics[width=1\linewidth]{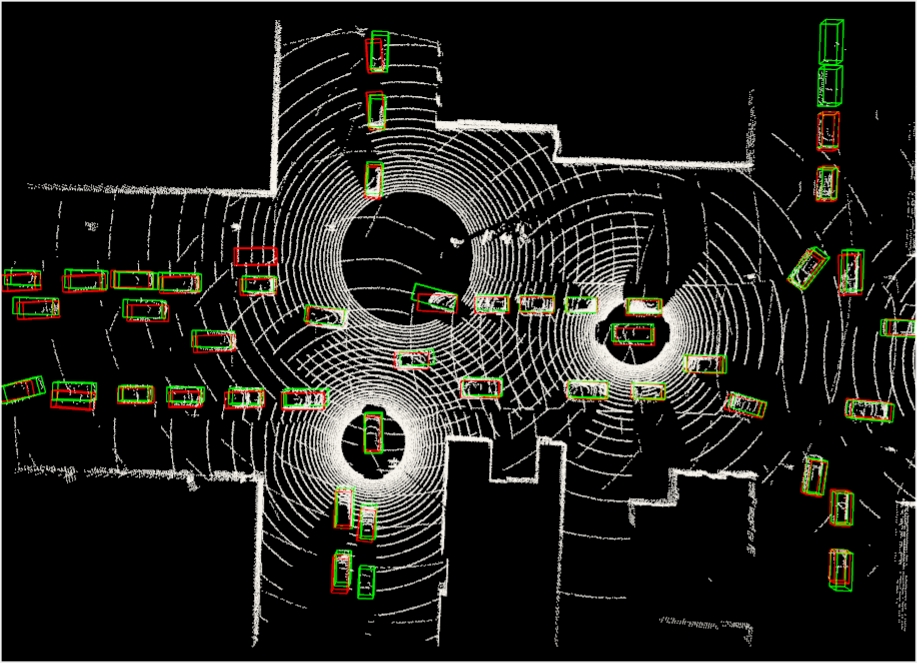}}
        \centering{\includegraphics[width=1\linewidth]{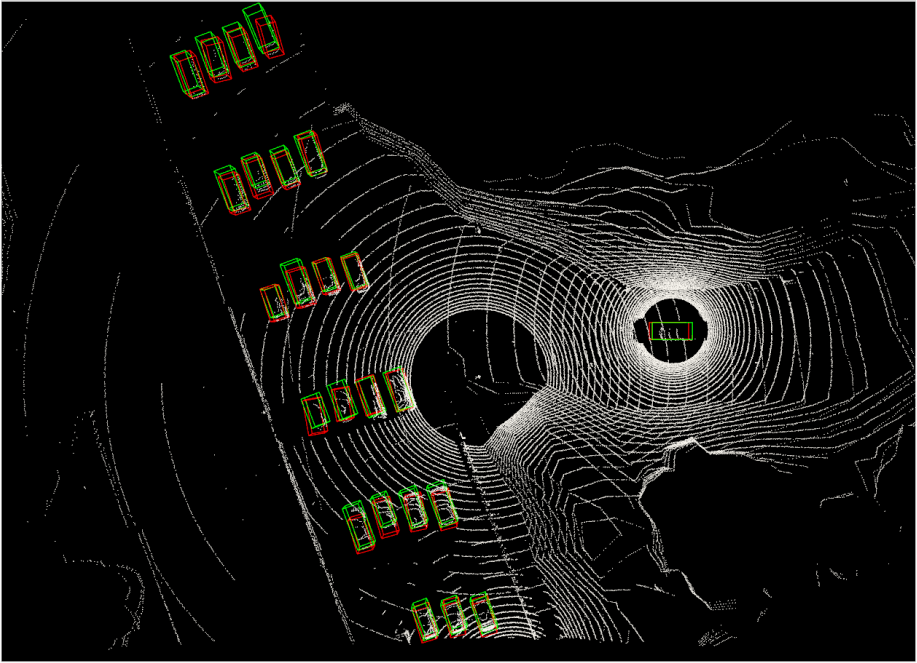}}
        \caption{DiscoNet~\cite{li2021learning}}
        \label{fig:output-c}
    \end{subfigure}
    \begin{subfigure}[c]{0.24\linewidth}
        \centering{\includegraphics[width=1\linewidth]{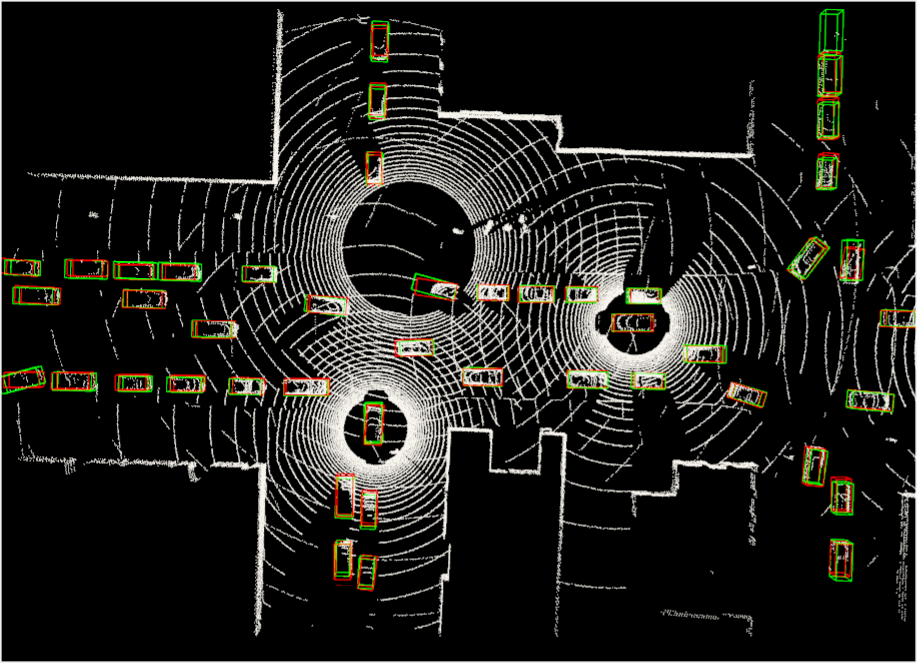}}
        \centering{\includegraphics[width=1\linewidth]{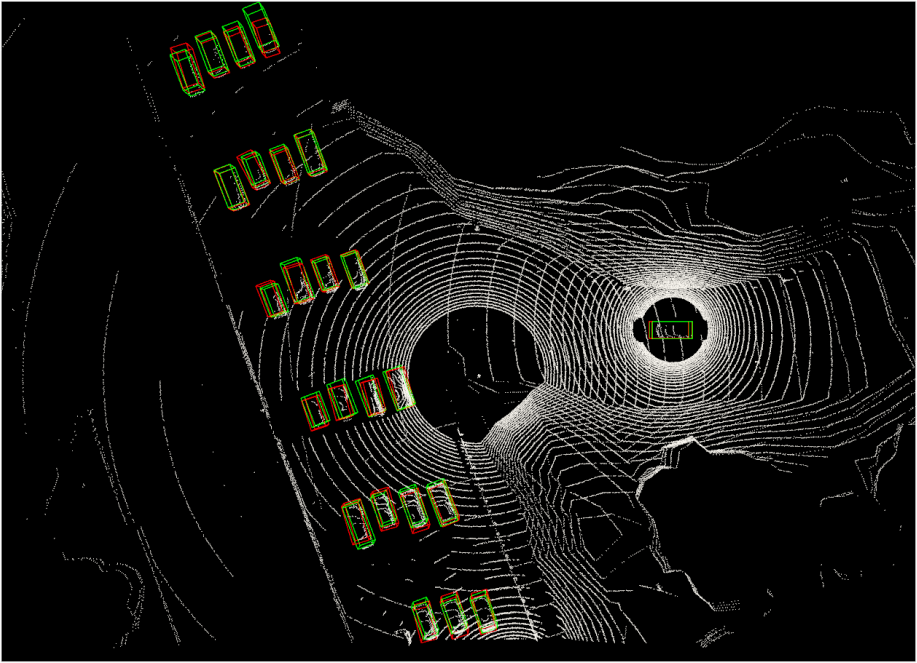}}
        \caption{V2X-ViT~\cite{xu2022v2xvit}}
        \label{fig:output-d}
    \end{subfigure}
    \caption{\textbf{Qualitative comparison in a congested intersection and a highway entrance ramp in V2XSet.} \textcolor{green}{Green} and \textcolor{red}{red} 3D bounding boxes represent the ground truth and prediction, respectively.}
    \label{fig:qualitive}
\end{figure*}

\subsection{Data}
\label{sec:data_management}
Data is an indispensable part of the OpenCDA. Especially for the perception system and learning-based decision-making/control, the models require a large-scale dataset for offline training in OpenCDA. Supported by the CDA data manager of OpenCDA, we have released two simulated large-scale datasets for cooperative perception. We also sent two L4 automated vehicles together to collect real-world data and develop a data acquisition and processing platform~\cite{xia2022automated} to facilitate field experiments for CDA.

\subsubsection{Simulated Dataset}
By utilizing OpenCDA and CARLA, we collect the first large-scale simulated V2V perception dataset OPV2V~\cite{xu2022opv2v}~(see figure~\ref{fig:data_sample}a for an example). OPV2V contains 11,464 frames of 3D LiDAR point clouds and RGB images from multiple vehicles in the same scenario with 230K annotated 3D
boxes. It supports multiple cooperative perception tasks, including cooperative detection, segmentation, and tracking. Since OPV2V was released, it has served as the foundation of cooperative perception~\cite{xu2022v2xvit,xu2022cobevt,hu2022where2comm,jiang2022map,lu2022robust}. Since OPV2V is a major focus on Vehicle-to-Vehicle cooperation, we proposed another dataset named V2XSet~\cite{xu2022v2xvit}, which adds the roadside units on OPV2V to enable Vehicle-to-Infrastructure cooperation. A data sample of V2XSet can be found in figure~\ref{fig:data_sample}b. 

\subsubsection{Real-world Dataset}
To accelerate the process of CDA research moving from simulation to the real world, we proposed an automated driving system (ADS) data acquisition and analytics platform (ADAAP) for vehicle trajectory extraction, reconstruction, and evaluation based on the connected automated vehicle (CAV) cooperative perception~\cite{xia2022automated}. This platform presents a holistic pipeline from the raw advanced sensory data collection to data processing, which can process the sensor data from multi-CAVs and extract the objects’ Identity (ID) number, position, speed, and orientation information in the map and Frenet coordinates. Besides the dynamic object information, this platform can also help users to generate HDMap directly from raw point clouds. Employing this platform, we sent out two CAVs and collected a real-world cooperative perception dataset that includes LiDAR point clouds, RGB camera images, annotated 3D bounding boxes, and HDMaps. A good sample of the collected dataset can be seen in figure~\ref{fig:data_sample}c. \rx{More importantly, our OpenCDA simulation tool is able to create a digital twin for the real-world dataset given the object annotation and xodr format map file. An example is shown in Fig.~\ref{fig:digit_twin}, where we load a real-world scenario into CARLA.} The data acquisition platform, along with the dataset, will be open-sourced as part of the ecosystem soon.

\subsection{Scenario Database}
OpenCDA offers a scenario database, generation, and
management tools, including various rule-based standard scenarios for testing different cooperative driving
applications as benchmarks. These
diverse scenarios will be used to evaluate the safety of driving systems in simulation. Since human-defined scenarios can only cover limited cases, we also develop an adversarial scene generator V2XP-ASG~\cite{xiang2022v2xp} to automatically infinitely generate challenging corner cases. Users can easily
replace any default modules in the OpenCDA with their designs and test them in the supplied or automatically
generated scenarios. If users desire to produce their scenarios, our framework also provides simple APIs to
support such customization.
\subsubsection{Rule-based Scenarios}
Our standard scenario database contains a series of YAML files that defines the specific scenario's parameters, including weather, traffic states, testing locations, etc. This database can be loaded by the OpenCDA simulation tool's scenario manager and any other platforms that follow the API format. The current scenario database includes classic scenes for platooning, cooperative merge, intersection control, etc.

\subsubsection{Adversarial Scenarios}
Rule-based scenarios only consider the common scenes and lack the capability of performing stress t ests on corner cases for the target perception systems. Thus, we introduce V2XP-ASG~\cite{xiang2022v2xp} – the first open adversarial scene generator for LiDAR-based V2X perception systems.  Unlike rule-based scenarios, V2XP-ASG can automatically generate abundant challenging scenes in an adversarial manner. V2XP-ASG first
constructs an adversarial collaboration graph by searching for
adversarial collaborators whose viewpoints combination will
lead to inferior performance. Afterward, it perturbs multiple
agents’ poses in an adversarial and plausible way. To reflect
the updated LiDAR observation caused by sensor viewpoint
change and pose perturbation, the high-fidelity CARLA [20] simulator is used to render the sensor measurements. Extensive
experiments show that V2XP-ASG can create challenging
scenes to severely deteriorate the performance of modern
V2X perception systems. More importantly, training in
these adversarial scenes can increase the system’s accuracy.

\subsection{Community}
\subsubsection{Ecosystem Growth}
\rx{OpenCDA aims to establish a community of practice (CoP) made up of various stakeholders in the industry, including developers of automated driving systems (ADS), OEMs, infrastructure technology developers, and universities and research institutes. The CoP has been increasing in its size since OpenCDA was released. The CoP provides feedback that will be incorporated into the OSE roadmap, and we will host webinars, panel discussions, and workshops to invite industry leaders, researchers, and government technical managers to share their insights. The OSE is initially based on our toolchains, but users can also create subsidiary software tools to perform specific tasks. Within the diverse backgrounds of the stakeholders, the roadmap can make the ecosystem easily support various customizations for different groups of people. To encourage innovation and growth within the  ecosystem, we will also hold a series of challenges for users to develop automated systems using the  platform and open their code. These challenges, called the OpenCDA Automated Driving Challenge, will be held annually or bi-annually in collaboration with conferences such as ICRA, IROS, ITSC, and CVPR. Participants will be given reproducible, diverse scenarios to test their systems on and will be evaluated based on metrics such as pass ratio and collision rate. The challenge will have two tracks: a full-stack system challenge and a specific component challenge. The topics for these tracks will be based on the updated roadmap, which is developed through stakeholder engagement in the CoP}.

\subsubsection{Community Building}
\rx{The OpenCDA ecosystem also engages with a group of early adopters called the Community of Content Contributors (CCC) to help develop and maintain the  platform. The CCC consists of researchers and developers in the AV/intelligent transportation industry and academia who have expressed interest in using and contributing to the platform and have done related work. We aim to hold monthly webinars to discuss the needs and priorities of the CCC and develop strategies to attract additional content contributors. In the beginning, our core development team is dedicated to providing start kits and tutorials for new users and contributors and build introductory materials in Python and Google Colab to lower the entry barrier for participation. These strategies for community building will be detailed in a CCC memorandum.}

\subsubsection{Governance and Organization}
\rx{OpenCDA will determine an appropriate governance model through discussions with COP and CCC. Many open-source governance models exist, including founder-leader, self-appointing council, electoral, and foundation-backed models. The most suitable one will be selected based on factors such as flexibility, simplicity, and scalability, as well as considerations like the proprietary nature of AV industry technologies and the availability of resources like vehicle testing and data collection. The chosen model should also support continuous development, integration, and deployment processes and infrastructure for open, asynchronous, and distributed development of the open-source product and ensure sustainability goals for the OSE. Currently, at the beginning stage of the ecosystem, we adopted the founder-led model, meaning that our core development team at UCLA takes the lead. The goal is to move away from a founder-led model to a community-managed approach for sustainability and scalability, although the foundation-backed model will also be explored. The governance model will also address development methodology, quality control, security and privacy, support for users, and onboarding mechanisms for new contributors.}

\begin{figure}
  \begin{center}
    \includegraphics[width=0.50\textwidth]{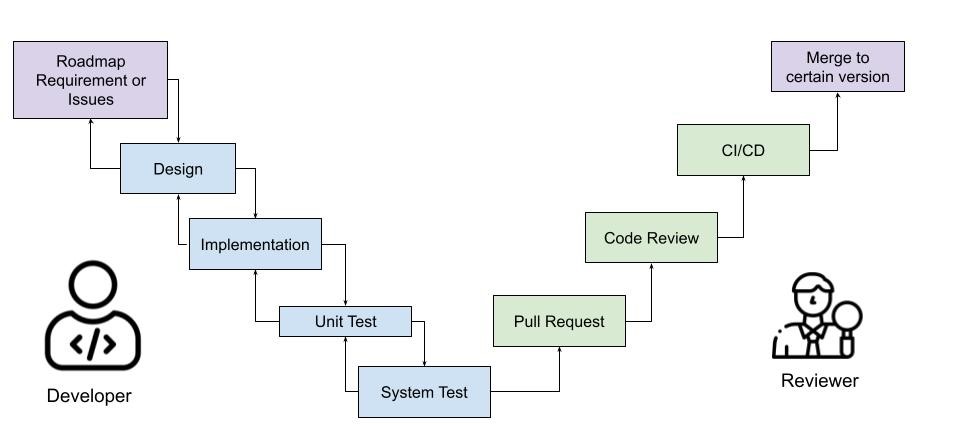}
  \end{center}
  \caption{The  healthy and self-sustainable life cycle of software development for the ecosystem.}
  \label{fig:software}
\end{figure}

\paragraph{Continuous Development and Quality Control.}\rx{ We aim to adopt a sustainable software development process that supports itself in the long run. Figure~\ref{fig:software} demonstrates the life cycle of the software development for ecosystem. As a standard process, the life cycle consists of two essential stages: the development stage and the review stage. In the development stage, a developer from the CCC will be assigned to a task based on the desired functionality from the roadmap or an issue raised by a user. The developer will design a pipeline, implement the code, write unit tests, and integrate the new functionality with the rest of the system. The developer will then create a pull request to move to the review stage. In the review stage, a qualified reviewer will statically and dynamically analyze the code and request any necessary improvements. If the pull request is approved, it will be merged into the correct branch/version of the codebase after passing an online CI/CD build. Eventually, the reviewer will merge this pull request to the correct branch/version of the codebase.}


\section{Case Studies}
\label{sec:4}

This section will conduct several case studies for different components of our OpenCDA ecosystem to demonstrate examples of advanced CDA research that can be enabled by OpenCDA. They include cooperative 3D LiDAR detection, cooperative merging with the classic modular pipeline, camera-based cooperative perception, and adversarial scene generation for the V2X perception system.  

\subsection{Case Study 1: Cooperative 3D LiDAR Detection}
\noindent\textbf{Scope.} In this case study, we demonstrate how OpenCDA enables cooperative LiDAR detection research for cooperative driving. We employ OpenCOOD to perform the cooperative 3D LiDAR detection task on \rx{three} datasets from the OpenCDA ecosystem, OPV2V, V2XSet, and \rx{paritial of the real-world dataset}. \rx{The real-world dataset contains  
14,000/2,000/4,000 frames for train/validation/test set, respectively. We ensure the real-world dataset has the same format as the simulated dataset to be compatible with OpenCOOD.} This requires models to be able to leverage multiple LiDAR views from different vehicles/infrastructures to perform 3D object detection on the ego vehicle. 

\noindent\textbf{Evaluation.} Th evaluation range in $x$ and $y$ direction are $[-140, 140]$~m and $[-40, 40]$~m respectively. The detection performance is measured with Average Precisions (AP) at Intersection-over-Union (IoU) thresholds of 0.5 and  0.7. In this study, we focus on LiDAR-based vehicle detection. Vehicles hit by at least one LiDAR point from any connected agent will be included as evaluation targets.

\noindent\textbf{Implementation details.}
During training, a random AV is selected as the ego vehicle, while during testing, we evaluate a fixed ego vehicle for all the compared models.  The communication range of each agent is set as 70~m based on, whereas all the agents out of this broadcasting radius of ego vehicle are ignored. We select PointPillar~\cite{lang2019pointpillars} as the backbone to convert raw LiDAR point clouds to 2D features.

\noindent\textbf{Compared models.} From our model zoo, we select SOTA multi-agent fusion models \textit{No Fusion}~(baseline), \textit{Early Fusion}, \textit{Late Fusion}, AttFuse~\cite{xu2022opv2v}, DiscoNet~\cite{li2021learning}, V2VNet~\cite{wang2020v2vnet}, F-Cooper~\cite{chen2019f}, and V2X-ViT~\cite{xu2022v2xvit} to perform the task.

\begin{table}[!t]
\centering
\scriptsize 
\setlength{\tabcolsep}{6pt}
\renewcommand{\arraystretch}{1.0}
\caption{\textbf{3D detection performance comparison on OPV2V and V2XSet.} We show Average Precision (AP) at IoU={0.5, 0.7}.}
\label{table:detection}
\begin{tabular}{lcccccc}

\cellcolor{lightgray} & \multicolumn{2}{c}{\cellcolor{lightgray} OPV2V}& \multicolumn{2}{c}{\cellcolor{lightgray} V2XSet} & \multicolumn{2}{c}{\cellcolor{lightgray} \rx{Real-world}}\\ 
\cline{2-3}\cline{4-5}\cline{6-7}
{\cellcolor{lightgray} Models}
& \cellcolor{lightgray} AP0.5& \cellcolor{lightgray} AP0.7& \cellcolor{lightgray} AP0.5& \cellcolor{lightgray} AP0.7  & \cellcolor{lightgray} AP0.5& \cellcolor{lightgray} AP0.7\\ \toprule
No Fusion           & 0.679          & 0.602          & 0.606          & 0.402   &  \rx{0.287} & \rx{0.134}       \\ 
Late Fusion          & 0.858          & 0.781          & 0.549          & 0.307     & \rx{0.439} & \rx{0.219}     \\
Early Fusion         & 0.891          & 0.800          & 0.720          & 0.384    & \rx{0.482} &   \rx{0.267}    \\
F-Cooper~\cite{chen2019f}             & 0.887          & 0.790          & 0.715          & 0.469       & \rx{0.456} & \rx{0.241}    \\
AttFuse~\cite{xu2022opv2v}                & \textbf{0.908}          & 0.815          & 0.709          & 0.487     & \rx{0.479} &   \rx{0.272}   \\
V2VNet~\cite{wang2020v2vnet}                & 0.897          & 0.822          & 0.791        & 0.493     & \rx{0.470} &  \rx{0.258}    \\
DiscoNet~\cite{li2021learning}& 0.899 & \textbf{0.836} & 0.798& 0.541 &\rx{ 0.468} & \rx{0.253} \\ 

V2X-ViT~\cite{xu2022v2xvit}          & 0.891 & 0.826 & \textbf{0.836} & \textbf{0.614} & \textbf{\rx{0.489}} & \rx{\textbf{0.295}} \\ 
\bottomrule
\end{tabular}
\end{table}

\noindent\textbf{Results.} Table~\ref{table:detection} shows the detection performance of different models on the \rx{three} cooperative perception datasets. It is clear that all the models that leverage the collaborations between agents outperform the single-vehicle perception baseline \textit{No Fusion} by a large margin on both datasets. In the OPV2V dataset, DiscoNet~\cite{li2021learning} achieves the best performance, slightly higher than the second best method V2x-ViT~\cite{xu2022v2xvit} by $1.0\%$ on AP@0.5. While \rx{in the V2XSet and real-world dataset}, V2X-ViT is the leading algorithm, outperforming DiscoNet by a large extent~($7\%$ and \rx{$4.2\%$}). This is reasonable as V2X-ViT has specific modules to handle GPS error and vehicle-infrastructure sensor heterogeneity, which only exist in the V2XSet and \rx{real-world dataset}. Figure~\ref{fig:qualitive} further demonstrates the qualitative results of selected SOTA models on two different scenarios in V2XSet. These results show how OpenCOOD can enable research studies on cooperative perception.

\begin{figure}[htbp]
\centering
 \includegraphics[width=\linewidth]{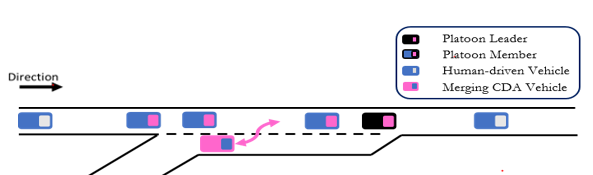} 
\caption{The cooperative merge testing scenario.}
\label{fig:cm}
\end{figure}

\begin{figure*}[htbp]
\centering
 \includegraphics[width=0.7\textwidth]{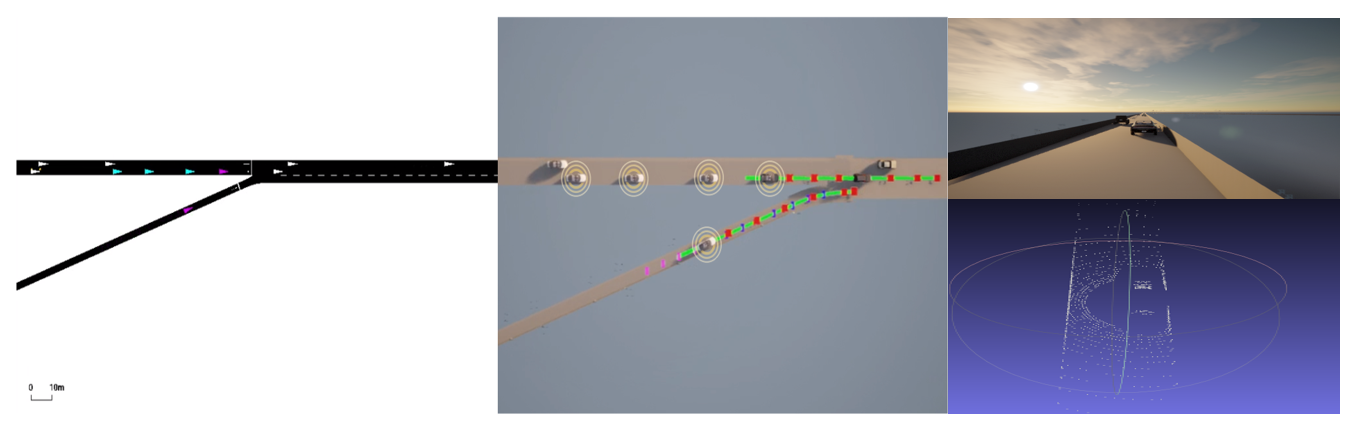} 
\caption{A snippet of cooperative merge scenario testing under co-simulation setting.}
\label{fig:cm2}
\end{figure*}

\begin{figure*}[!t]
\vspace{2mm}
\centering
    \begin{subfigure}[c]{0.21\linewidth}
        \centering{\includegraphics[width=1\linewidth]{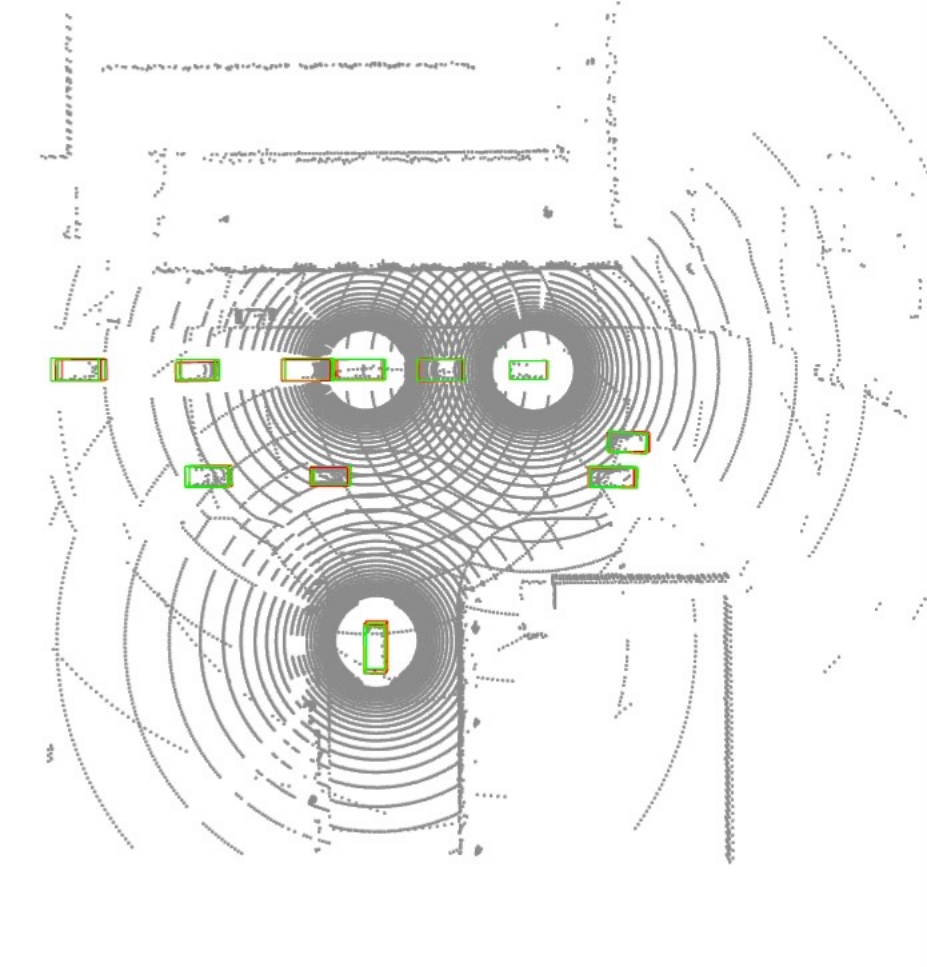}}
        \caption{Initial scene}
        \label{fig:qualitive-a}
    \end{subfigure}
    \begin{subfigure}[c]{0.21\linewidth}
        \centering{\includegraphics[width=1\linewidth]{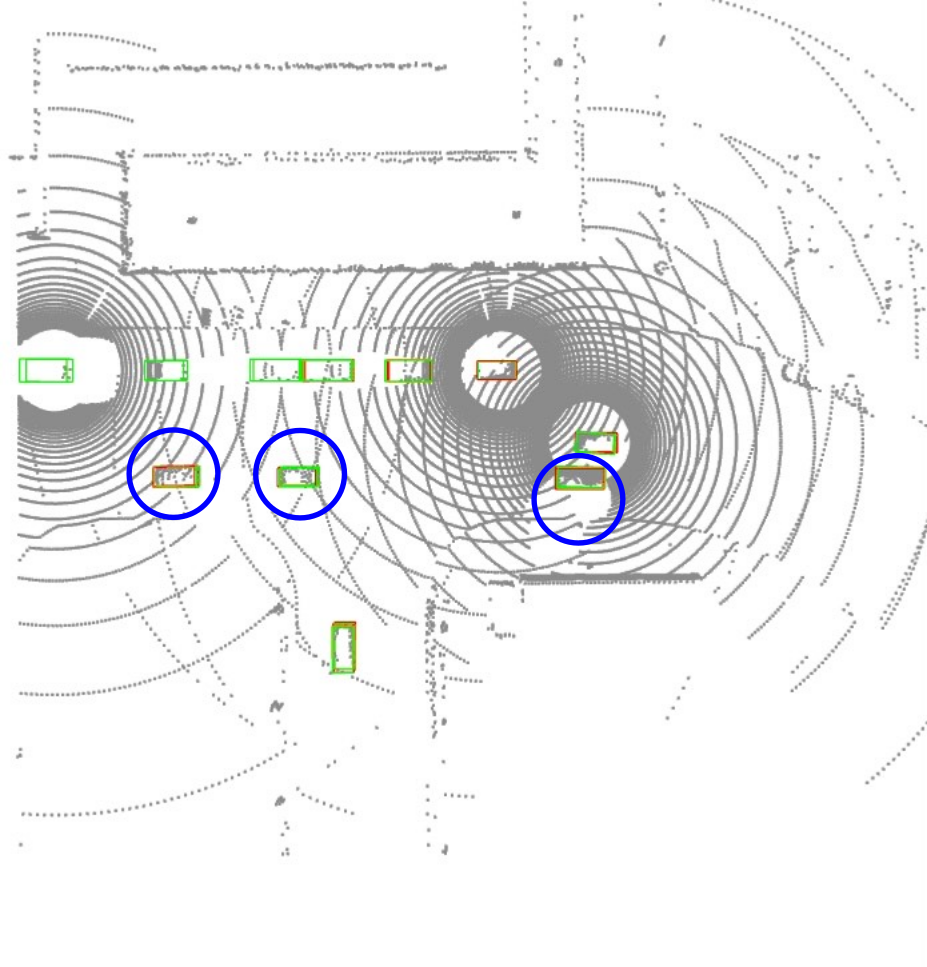}}
        \caption{After ACS}
        \label{fig:qualitive-b}
    \end{subfigure}
    \begin{subfigure}[c]{0.21\linewidth}
        \centering{\includegraphics[width=1\linewidth]{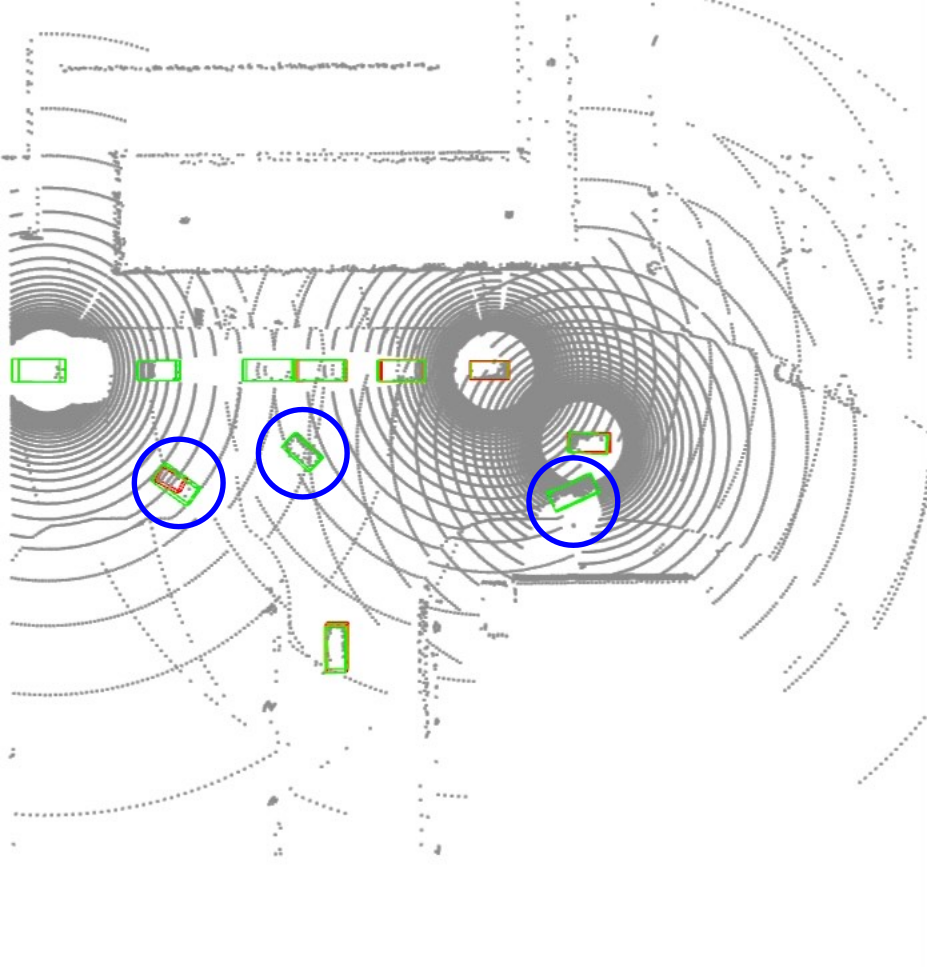}}
        \caption{Final scene}
        \label{fig:qualitive-c}
    \end{subfigure}
    \caption{Qualitative visualization of  V2XP-ASG generated challenging scenes. (a) Initial scene. (b) Scene after adversarial collaborator search. (c) Scene after adversarial pose perturbation.  Here blue circles represent the perturbed vehicles.}
    \label{fig:qualitive}
    \vspace{-3mm}
\end{figure*}

\subsection{Case Study 2: Cooperative Merge into Platoons}
\noindent\textbf{Scope.} In this case study, we demonstrate how OpenCDA enables researchers to flexibly use the co-simulation functionality to flexibly develop and test cooperative decision making and control algorithms. We conduct experiments for cooperative merge using the OpenCDA simulation tool (co-simulation functionality). The scenario design is shown in figure~\ref{fig:cm}. The mainline has a high-speed traffic flow
mixed with human-driven vehicles managed by SUMO and
CAVs that are controlled by our own cooperative driving systems. When the single CAV is near
the merging area, it will communicate with the mainline
platoon and request to join. Once they agree, the single CAV has to finish the merge and join the platoon simultaneously before the acceleration lane
ends. The leader will decide the best merging position and
command certain platoon members to create a gap for the
new member. Figure~\ref{fig:cm2} shows a snippet of the platooning co-simulation testing using a customized benchmark map of a basic freeway merge segment included in the OpenCDA simulation tool.

\noindent\textbf{Evaluation.} The whole system is evaluated using safety, stability, and efficiency.
\begin{itemize}
    \item \textbf{Safety.} We evaluate the safety using two indicators, Time-to-Collision~(TTC) and hazard frequency. TTC refers to the time required for two vehicles to collide if they continue at their present speed and on the same path. Hazard frequency is defined as the number of events that $TTC < TTC_t$, where $TTC_t$ is the warning threshold of $TTC$ to distinguish between safe and unsafe events. We set the threshold as $2.5$ seconds.
    \item \textbf{Stability.} The stability of a platoon refers to whether oscillations are amplified from downstream to upstream vehicles. Two metrics are used to measure the stability: inter-vehicular time gap and accelerations. The inter-vehicular time gap is the time difference between a platoon member and the preceding member. The time-series data of acceleration and statistics of the data, such as mean and standard deviation, are calculated to indicate the driving smoothness of the platoon members.
    \item \textbf{Efficiency.} The efficiency of platoon joining is measured by the time duration required for the maneuver and the smoothness of the process. The time it takes to complete the maneuver, and the acceleration during the joining process are used as indicators of efficiency.
\end{itemize}

\noindent\textbf{Implementation details.} In this testing scenario, there are a total of 5 connected and automated vehicles (CAVs), including four in the platoon and one on the entrance ramp. All CAVs have the classic modular driving pipeline, including perception, localization, planning, and control. For perception, Yolov5~\cite{glenn_jocher_2022_7002879,liu2022yolov5} with simple late LiDAR fusion is used for 3D object detection. For localization, Kalman Filter with GPS and IMU is used, and PID controller is used for control. In terms of planning, the leading vehicle of the platoon uses a normal rule-based trajectory planning, while the others use the Platooning finite state machines algorithm~\cite{han2022strategic}.

\noindent\textbf{Compared models.} A critical decision for cooperative merging is the merging position. Here we compare two different algorithms: the heuristic-based method and Genetic Fuzzy System~(GFS)~\cite{Cordon2001}. The first approach  will choose the vehicle in the platoon
that has the shortest Euclidean distance as the frontal vehicle
for merging. The second one utilizes fuzzy logic to decide the best merging position. Different from a heuristic-based method, it also
takes platoon members’ speed and surrounding human-driven
vehicles’ information into consideration.

\noindent\textbf{Results.} Table~\ref{tab:results} reveals the cooperative merging results with two different algorithms. Both algorithms achieve good safety and stability with high average TTC, 0 hazard frequency, and slight acceleration fluctuations for platooning. In terms of efficiency, the GFS method performs favorably against the heuristic method.

\begin{table}[]
\begin{subtable}[h]{0.45\textwidth}
\begin{tabular}{l|ll|lll|ll}
\cline{1-6}
\cellcolor[HTML]{C0C0C0} &
  \multicolumn{2}{c|}{\cellcolor[HTML]{C0C0C0}Safety} &
  \multicolumn{3}{l|}{\cellcolor[HTML]{C0C0C0}{\color[HTML]{000000} Stability}} &
  \multicolumn{2}{l}{\cellcolor[HTML]{C0C0C0}Efficiency} \\ \cline{2-8} 
\multirow{-2}{*}{\cellcolor[HTML]{C0C0C0}Vehicle id} &
  attc &
  hf &
  atg &
  tg\_std &
  acc\_std &
  tcm &
  acc\_std \\ \hline
0 & 49.8    & 0 & NA    & NA    & 0.92 & NA & 0.03 \\ \hline
1 & 25.5 & 0 & 0.607 & 0.005 & 0.63 & NA & 0.01 \\ \hline
2 & 31.03 & 0 & 0.607 & 0.003 & 0.56 & NA & 0.01 \\ \hline
3 & 31.53 & 0 & 0.612 & 0.013 & 1.33 & 13.5 & 2.83 \\ \hline
4 & 32.59 & 0 & 0.707 & 0.23 & 0.82 & NA & 1.37 \\ \hline
\end{tabular}
\caption{Cooperative merge and platoon join using heuristic method}
\end{subtable}
\label{tab:results_d}

\begin{subtable}[h]{0.45\textwidth}
\begin{tabular}{l|ll|lll|ll}
\cline{1-6}
\cellcolor[HTML]{C0C0C0} &
  \multicolumn{2}{c|}{\cellcolor[HTML]{C0C0C0}Safety} &
  \multicolumn{3}{l|}{\cellcolor[HTML]{C0C0C0}{\color[HTML]{000000} Stability}} &
  \multicolumn{2}{l}{\cellcolor[HTML]{C0C0C0}Efficiency} \\ \cline{2-8} 
\multirow{-2}{*}{\cellcolor[HTML]{C0C0C0}Vehicle id} &
  attc &
  hf &
  atg &
  tg\_std &
  acc\_std &
  tcm &
  acc\_std \\ \hline
0 & 49.8    & 0 & NA    & NA    & 0.95 & NA & 0.02 \\ \hline
1 & 31.40 & 0 & 0.608 & 0.007 & 1.27 & 9.9 & 2.51 \\ \hline
2 & 31.24 & 0 & 0.674 & 0.16 & 0.79 & NA & 1.18 \\ \hline
3 & 30.14 & 0 & 0.609 & 0.008 & 0.65 & NA & 0.88 \\ \hline
4 & 29.9 & 0 & 0.607 & 0.006 & 0.61 & NA & 0.59 \\ \hline
\end{tabular}
\caption{Cooperative merge and platoon join using GFS}
\end{subtable}
\label{tab:results_e}

\caption{\textbf{Quantitative results of two different scenario tests}. \textmd{The desired platoon time gap is set to 0.6 seconds. attc:average time-to-collision(second), hf:hazard frequency(number of times), atg:average platoon time gap(second), tg\_std:platoon time gap standard deviation(second), tcm: time to complete maneuver(second)} }
\label{tab:results}
\end{table}

\subsection{\rx{Case Study 3: Camera-based Cooperative Perception}}
\noindent\textbf{Scope.} \rx{In this case study, we demonstrate how OpenCDA enables researchers to innovate in multi-modal sensing and perception research via the use our flexible framework and data processing structure. We aim to investigate the use of camera-based cooperative perception approaches for the task of bird's-eye-view (BEV) semantic segmentation~\cite{zhou2022cross,xu2022cobevt} in dynamic environments. To do this, we use both the OpenCDA simulation tool and OpenCOOD, focusing on the OPV2V dataset. Since the OPV2V dataset does not include map information, we use the driving log replay functionality in the OpenCDA CDA Data Loader to generate ground truth for the HDMap. Each connected vehicle in the case study has four cameras facing in different directions. To perform the BEV segmentation, the models must first infer 3D information from the multi-view 2D images and then aggregate the inferred 3D information from all the vehicles to predict the BEV segmentation mask.}

\noindent\textbf{Evaluation.} \rx{During testing, we select a fixed vehicle as the ego vehicle and evaluate the performance of our models within a $100m \times 100m$ area around it with a map resolution of $39cm$. We use the intersection over union (IoU) between the predicted map and the ground truth map-view labels as our performance metric. We segment both dynamic objects and the topology of the map.}

\noindent\textbf{Compared models.} \rx{For the multi-agent perception task, we consider a single-agent perception system without fusion as the baseline. We compare the performance of this baseline system with state-of-the-art multi-agent perception algorithms such as F-Cooper~\cite{chen2019f}, AttFuse~\cite{xu2022opv2v}, V2VNet~\cite{wang2020v2vnet}, DiscoNet~\cite{li2021learning}, and CoBEVT~\cite{xu2022cobevt}. Additionally, we also implement a straightforward fusion strategy called Map Fusion, which transmits the segmentation map instead of BEV features and fuses all the maps by selecting the closest agent's prediction for each pixel.}

\noindent\textbf{Results.} \rx{As shown in  Tab.~\ref{tab:opv2vcamera}, all cooperative methods perform better than \textit{No Fusion}, which proves the benefits of multi-agent perception system. Among all fusion models,  CoBEVT achieves the best IoU for all classes, outperforming the second-best method by 7.5\%, 2.3\%, and 7.2\% on the vehicle, drivable area, and lane, respectively. We also show the qualitative results of CoBEVT on scenes containing 3 AVs in Fig.~\ref{fig:qualitive}. In each row, we draw
the front camera image of each AV, along with the ground truth and prediction pairs. Through collaboration, our framework can overcome most of the occlusions and perceive distant objects accurately, benefiting from our Transformer design that learns from all agents and views.}

\begin{figure*}[!t]
\centering
    \begin{subfigure}[c]{0.19\linewidth}
        \centering{\includegraphics[width=1\linewidth]{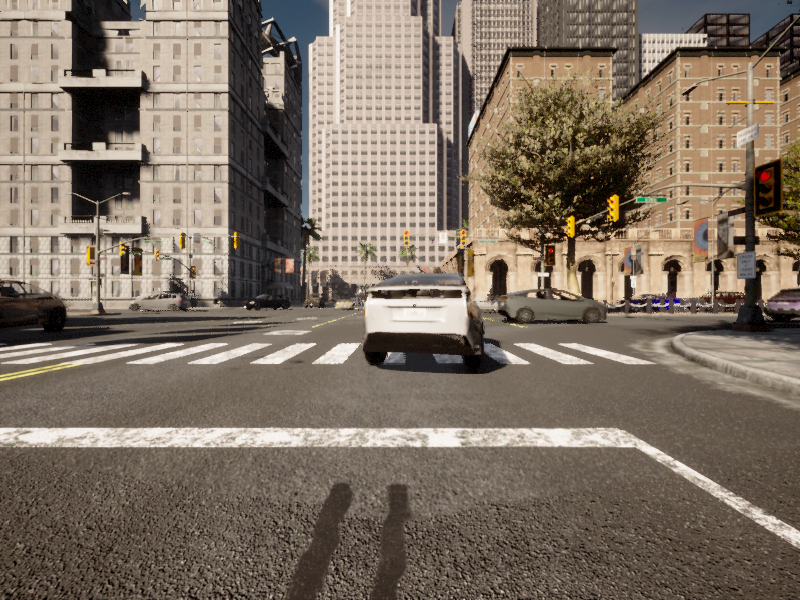}}
        \centering{\includegraphics[width=1\linewidth]{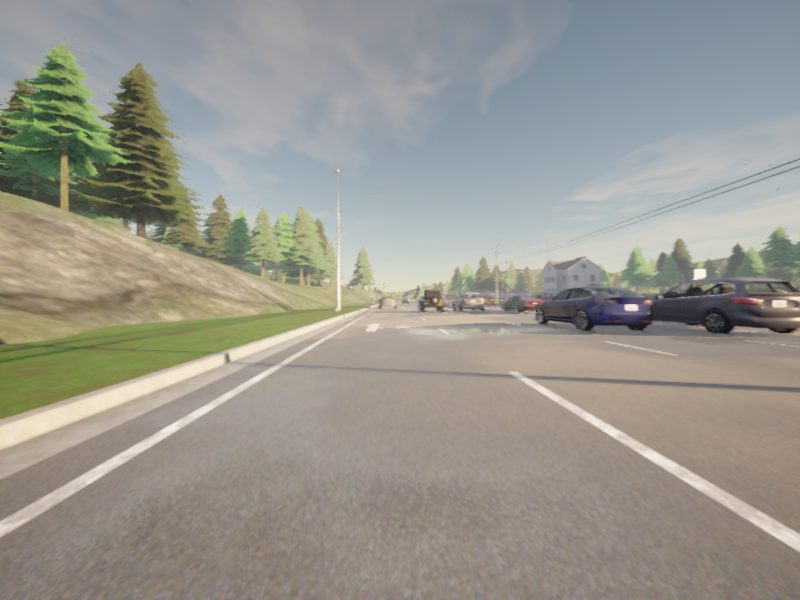}}
        \caption{ego}
        \label{fig:output-a}
    \end{subfigure}
    \begin{subfigure}[c]{0.19\linewidth}
        \centering{\includegraphics[width=1\linewidth]{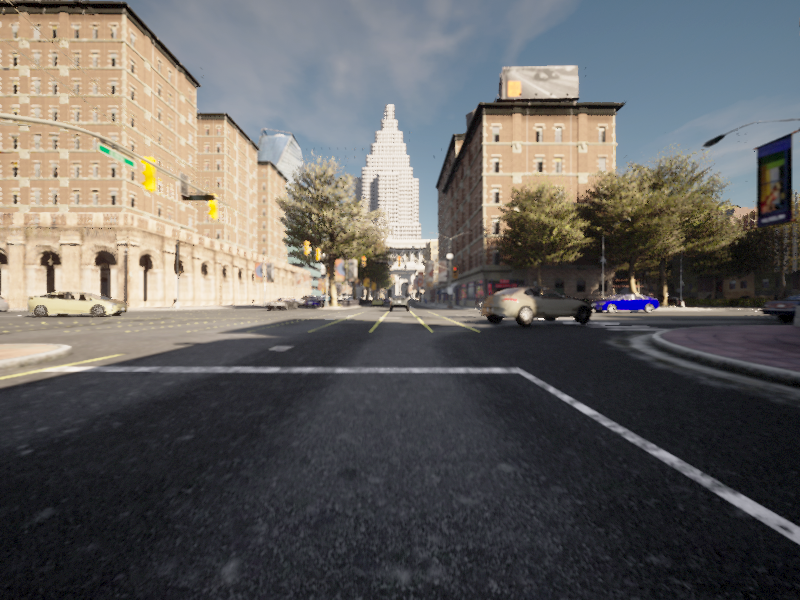}}
        \centering{\includegraphics[width=1\linewidth]{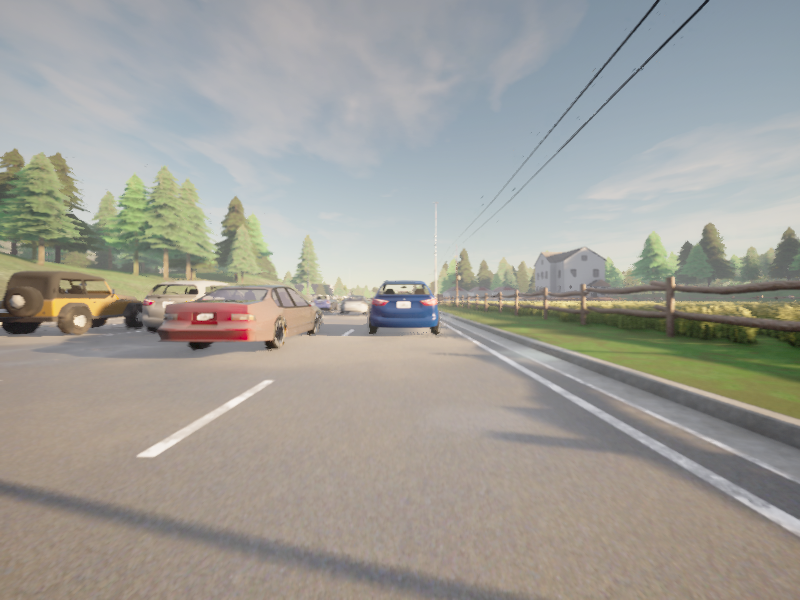}}
        \caption{av1}
        \label{fig:output-b}
    \end{subfigure}
    \begin{subfigure}[c]{0.19\linewidth}
        \centering{\includegraphics[width=1\linewidth]{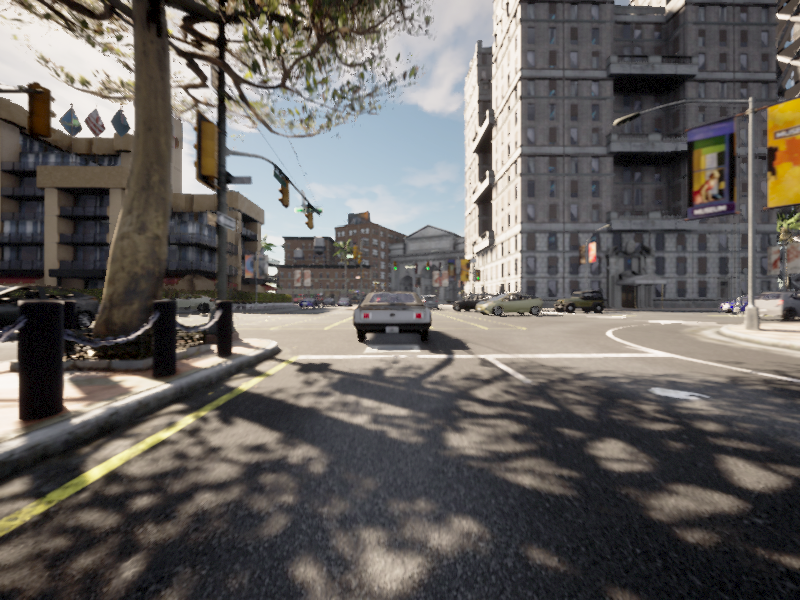}}
        \centering{\includegraphics[width=1\linewidth]{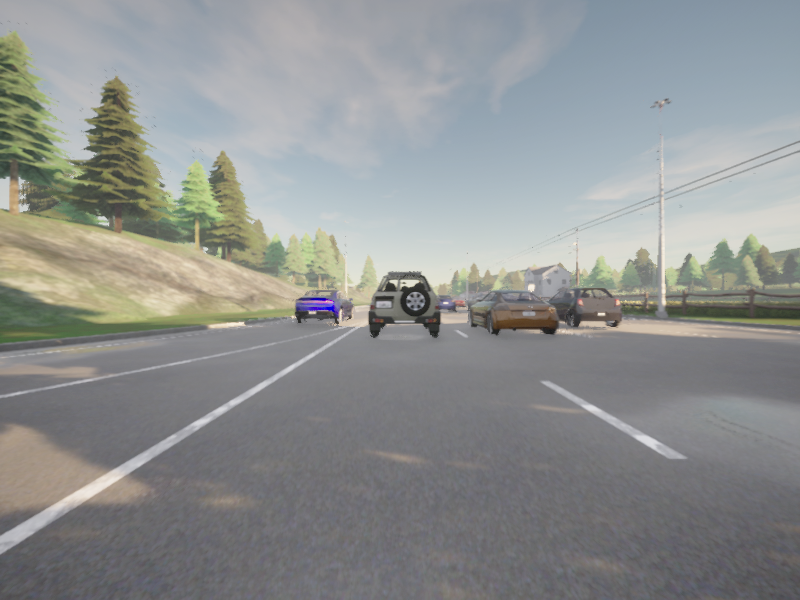}}
        \caption{av2}
        \label{fig:output-c}
    \end{subfigure}
    \begin{subfigure}[c]{0.19\linewidth}
        \centering{\includegraphics[width=1\linewidth]{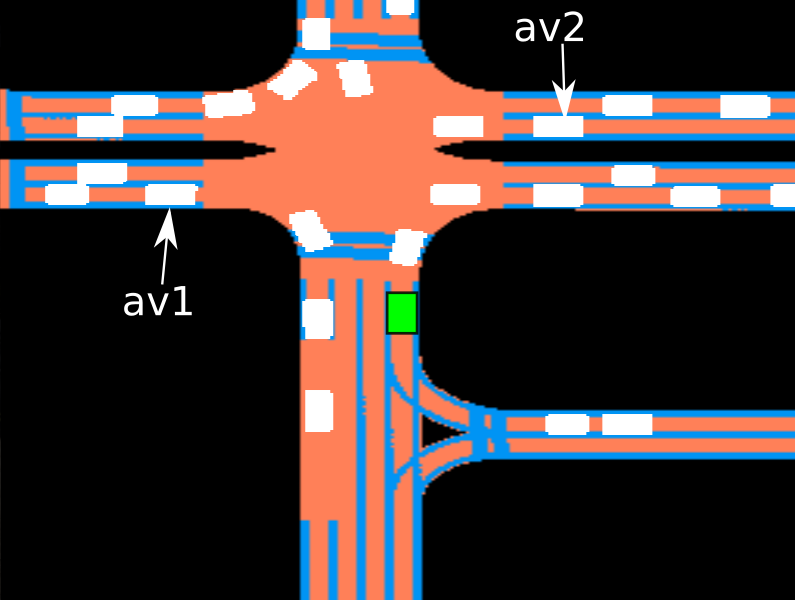}}
        \centering{\includegraphics[width=1\linewidth]{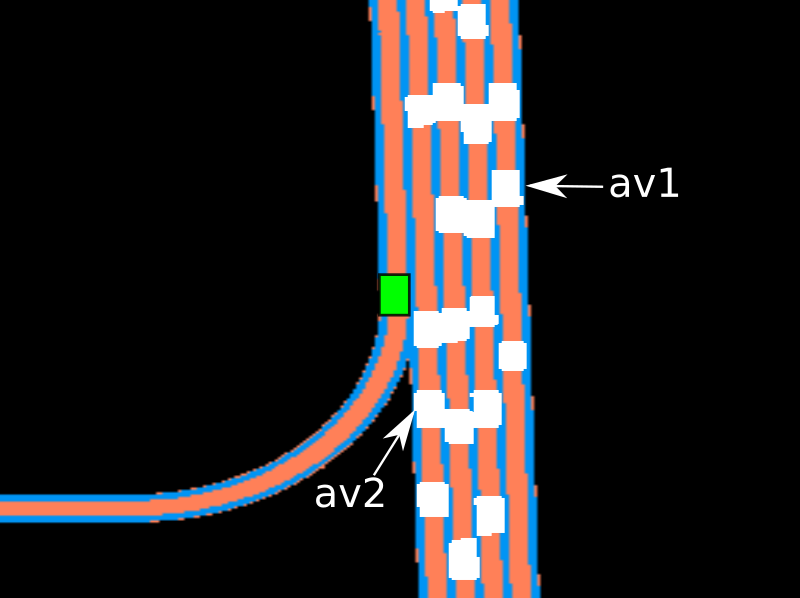}}
        \caption{groundtruth}
        \label{fig:output-d}
    \end{subfigure}
        \begin{subfigure}[c]{0.19\linewidth}
        \centering{\includegraphics[width=1\linewidth]{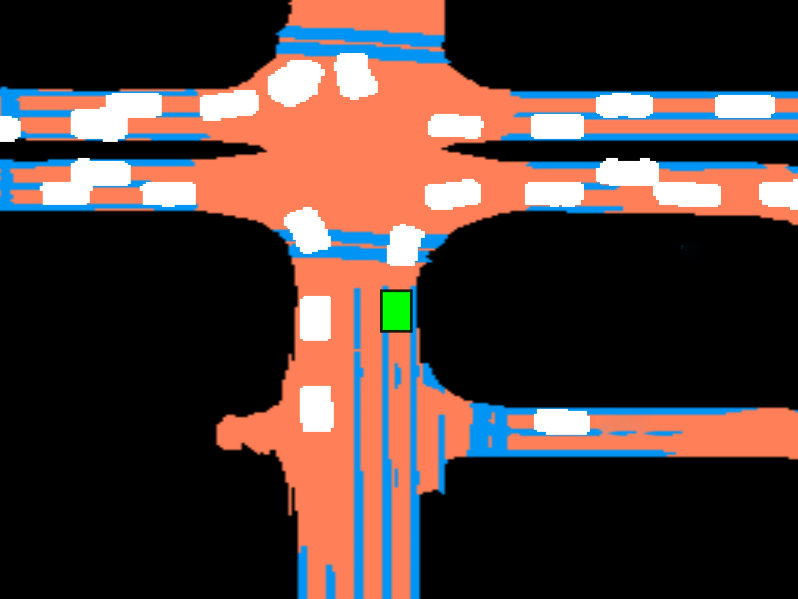}}
        \centering{\includegraphics[width=1\linewidth]{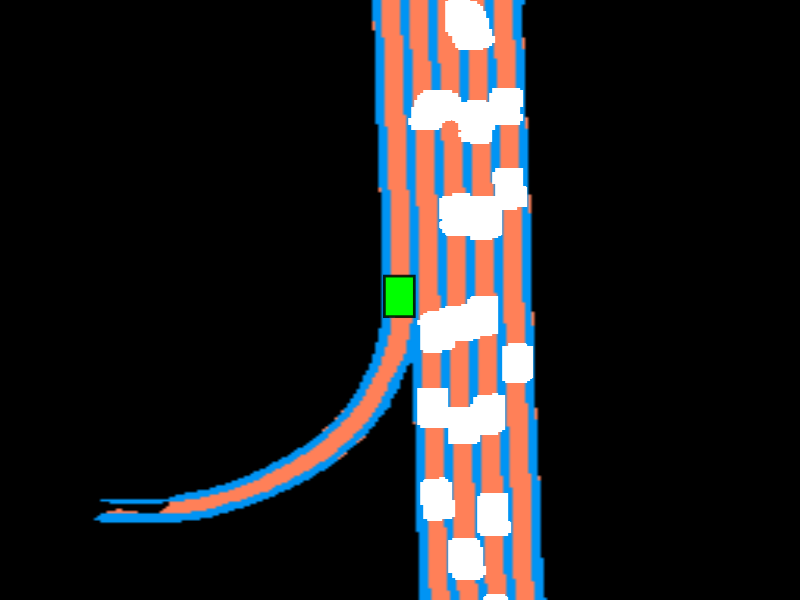}}
        \caption{prediction}
        \label{fig:output-e}
    \end{subfigure}
    \vspace{-2mm}
    \caption{\textbf{Qualitative results of case study 3.} \rx{From left to right: the front camera image of (a) ego, (b) av1, (c) av2, (d) groundtruth and (e) prediction of CoBEVT. The green bounding boxes represent ego vehicle, while the white boxes denote the segmented vehicles. CoBEVT demonstrates robust performance under various traffic situations and road types. }
    }
    \label{fig:qualitive}
    \vspace{-2mm}
\end{figure*}

\begin{table}
\centering
\footnotesize
\caption{\textbf{Map-view segmentation on OPV2V camera-track.} We report IoU for all classes.}
\label{tab:opv2vcamera}
\begin{tabular}{l|ccc}
\cellcolor{lightgray} Method & \cellcolor{lightgray}Veh. & \cellcolor{lightgray}Dr.Area & \cellcolor{lightgray}Lane\\
\toprule
No Fusion & 37.7 &57.8 & 43.7 \\\hline
 Map Fusion & 45.1 & 60.0 & 44.1 \\
 F-Cooper~\cite{chen2019f} & 52.5 & 60.4 & 46.5 \\
 AttFuse~\cite{xu2022opv2v} & 51.9 & 60.5 & 46.2 \\
  
 V2VNet~\cite{wang2020v2vnet} & 53.5 &60.2 & 47.5 \\  
 DiscoNet~\cite{li2021learning} & 52.9 & 60.7 & 45.8\\ \hline
 CoBEVT & \textbf{60.4} & \textbf{63.0} & \textbf{53.0}\\
\end{tabular}
\end{table}

\subsection{Case Study 4: Adversarial Scene Generation}
\noindent\textbf{Scope.} In this case study, we illustrate how OpenCDA can be utilized as a framework for researchers to generate challenging scenarios for testing and training their algorithms. We utilize one of our core team's algorithms, V2XP-ASG, to create challenging scenes for V2X perception systems using 331 scenes sampled from the augmented OPV2V dataset. We manipulate normal scenes in the database in an adversarial and realistic manner by modifying agent positions and collaborator choices. Our approach involves two stages: 1) adversarial collaborator search, where collaborators are probabilistically selected based on the attention weights of an intermediate fusion model, and 2) adversarial perturbation search, where we utilize historical observations of perturbations and loss pairs and employ black-box optimization algorithms to generate perturbations that can result in inferior performance. Additionally, we ensure the perturbations are physically plausible by eliminating perturbations that could cause collisions or positions that are out of range.

\noindent\textbf{Evaluation: }  We benchmark V2XP-ASG's performance for several cooperative perception algorithms, including No Fusion, Late Fusion, Early Fusion, and AttFuse~\cite{xu2022opv2v}. We conduct the experiments based on the CARLA simulator and evaluate the model for the range: $x,y\in[-48,48]$ m. For consistency, the ego agent is fixed during the whole optimization process and evaluation. To reduce the influence of noise on the perception performance, we set zero noise for LiDAR simulation so that the performance drop is solely dependent on the intrinsic scene configurations. 

\noindent\textbf{Results: }In Table.~\ref{tab:main_table}, we benchmarked the performance drop of V2XP-ASG generated challenging scenes compared with original normal scenes for various V2X fusion methods. From the high AP drop in the challenging scenes, we can see that V2XP-ASG can identify challenging scenes for diverse models with different fusion strategies.
In Figure.~\ref{fig:qualitive}, we demonstrate the qualitative results of generated challenging scenes. After adversarial collaborator search and adversarial perturbation search, several vehicles are mis-detected, which imposes a severe challenge for the modern V2X perception systems. 

\begin{table}[]
\vspace{1.6mm}
\footnotesize
\setlength{\tabcolsep}{2pt}
    \centering

    \begin{tabular}{c|c|l|l|l}
         \cellcolor[HTML]{C0C0C0} {Methods}&\cellcolor[HTML]{C0C0C0} {Scenes} \cellcolor[HTML]{C0C0C0} {Type}&\cellcolor[HTML]{C0C0C0} {AP@0.3}&\cellcolor[HTML]{C0C0C0} {AP@0.5}& \cellcolor[HTML]{C0C0C0} {AP@0.7}\\
         \toprule
         \multirow{2}{*}{No Fusion}&Normal&55.4 & 54.8 & 46.3\\
         &Challenging&31.1(\textbf{-24.3}) & 30.3(\textbf{-24.5}) & 25.1(\textbf{-21.2})\\
         \midrule
         \multirow{2}{*}{Late Fusion}&Normal&73.4 & 72.6 & 62.6\\
         &Challenging& 42.0(\textbf{-31.4}) & 40.0(\textbf{-32.6}) & 32.8(\textbf{-29.8})\\
         \midrule
         \multirow{2}{*}{Early Fusion}&Normal&80.3 & 79.8 & 73.5\\
         &Challenging&49.8(\textbf{-30.5}) & 48.3(\textbf{-31.5}) & 43.4(\textbf{-30.1})\\
         \midrule
         \multirow{2}{*}{AttFuse}&Normal&82.4&81.5& 74.6\\
         &Challenging & 46.6(\textbf{-35.8}) & 44.9(\textbf{-36.6}) & 40.5(\textbf{-34.1})\\
         \bottomrule
    \end{tabular}

    \caption{Evaluation of V2X perception models on initial 
    \textit{Normal} scenes and V2XP-ASG generated \textit{Challenging} scenes. The number in bold is the AP drop. }
        \label{tab:main_table}
    \vspace{-6mm}
\end{table}

\section{Conclusions}
\label{sec:5} 
In this paper, we introduce the OpenCDA ecosystem, which integrates a model zoo, a suite of driving simulators
at various resolutions, benchmark training/testing datasets,
and a scenario database/generator for CDA research. We describe how the ecosystem work and grows and conduct several case studies to demonstrate its capability of facilitating CDA research from different perspectives.

 \rx{In the future, we plan to continue expanding and improving our OpenCDA simulation tool by adding missing components such as a human interaction layer and end-to-end learning. We will also continually grow our model zoo to support more research studies. Additionally, we plan to interface with real systems in various ways. For example, the perception model trained using OpenCOOD can be transferred to real vehicles, and we will provide a corresponding toolchain to help users easily migrate the models for real-world deployment. We also plan to develop a ROS version of the OpenCDA simulation tool, which can interact with the real system in two ways: 1) it can be used to provide the environment for hardware-in-the-loop testing, and 2) the components of the cooperative driving system (e.g., perception, planning) in OpenCDA can be directly deployed on the real vehicle.}

\rx{The development of OpenCDA will also heavily rely on joint efforts from the community. As OpenCDA gradually transits from the founder-led model to a community-managed approach, contributions from the community to the ecosystem are vital for the future continuous development of OpenCDA. We strongly encourage existing users of OpenCDA and other CDA developers to contribute their work to the OpenCDA community.}




\ifCLASSOPTIONcaptionsoff
  \newpage
\fi

\bibliographystyle{IEEEtran}
\bibliography{reference}

%




\vspace{-2em}
\begin{IEEEbiography}
[{\includegraphics[width=1in,height=1.35in,clip,keepaspectratio]{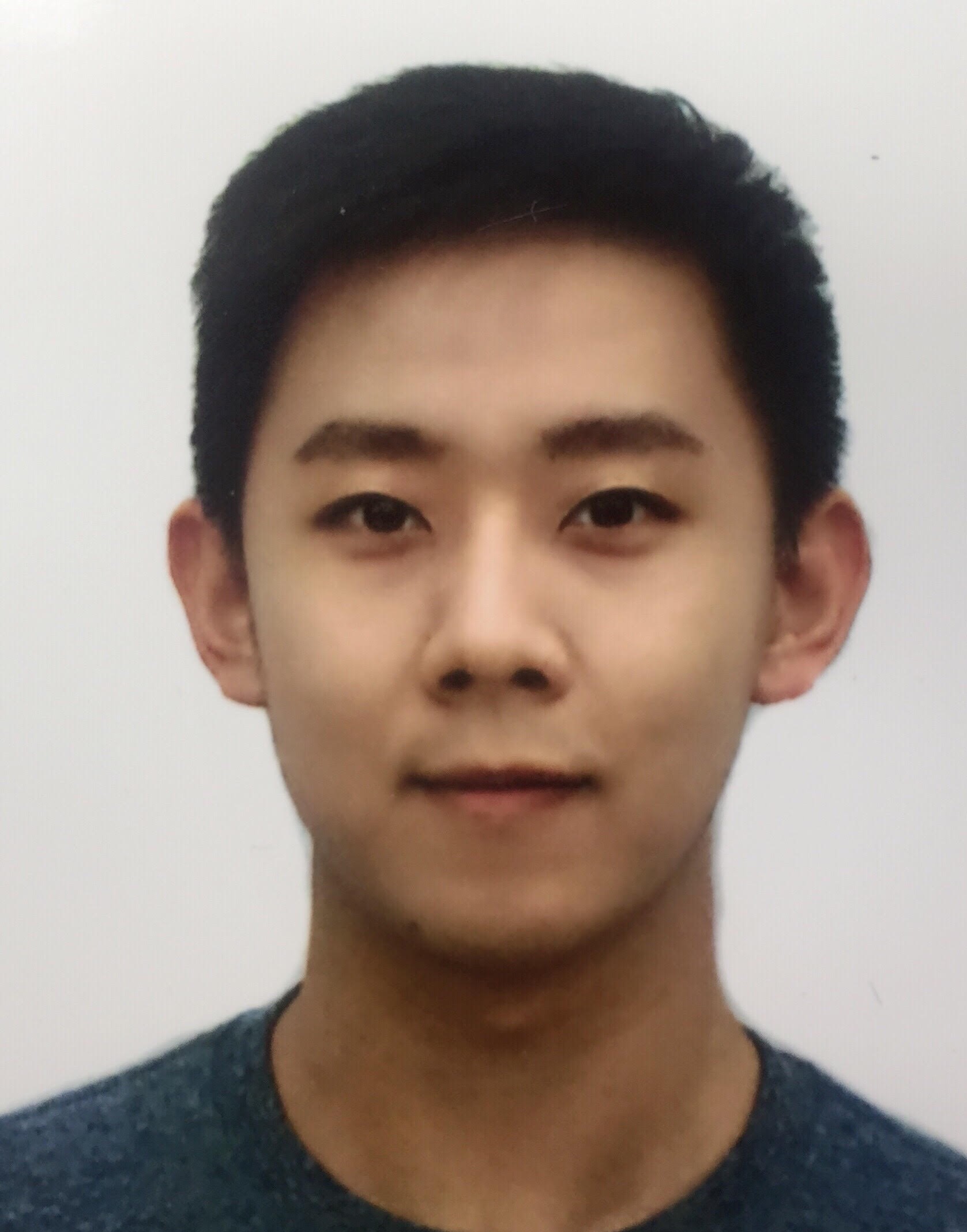}}]{Runsheng Xu} received the B.S. and M.S. degrees in Electrical Engineering from North China Electrical Power University, Beijing, China, and Northwestern University, Evanston, USA, in 2016 and 2017, respectively. He is currently pursuing the Ph.D. degree with the UCLA Mobility Lab,  University of California, Los Angeles. His research interests include computer vision, autonomous driving perception system, and cooperative driving automation.
\end{IEEEbiography}

\begin{IEEEbiography}[{\includegraphics[width=1in,height=1.25in,clip,keepaspectratio]{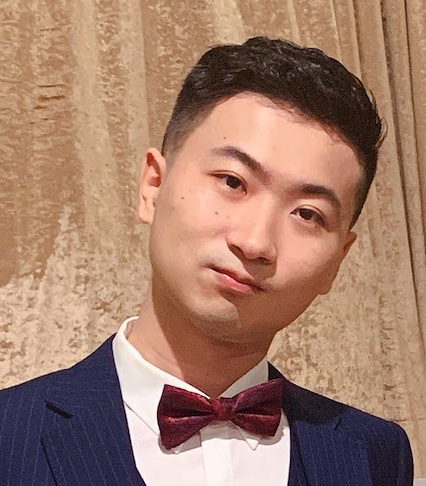}}]{Hao Xiang} received the B.S. degree in Microelectronics from Fudan University and M.S. degree in Machine Learning and Data Science from University of California, San Diego. He is currently a Ph.D. student at UCLA mobility Lab, University of California, Los Angeles. His research interests include computer vision, multi-agent learning, multi-modal learning, trajectory prediction, robotics, and autonomous driving.
\end{IEEEbiography} 

\begin{IEEEbiography}[{\includegraphics[width=1in,height=1.25in,clip,keepaspectratio]{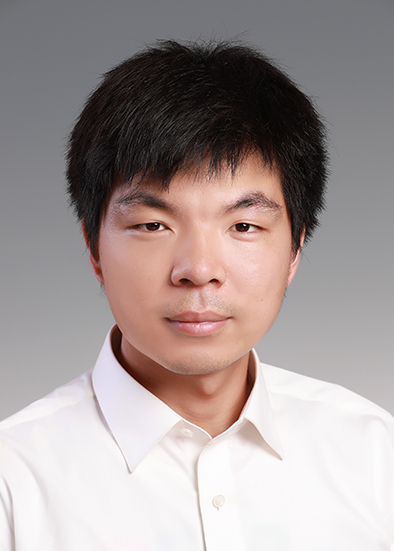}}] {Xin Xia} received the B.E. degree in vehicle engineering from the School of Mechanical and Automotive Studies, South China University of Technology, Guangzhou, China, in 2014, and the Ph.D. degree in vehicle engineering from the School of Automotive Studies, Tongji University, Shanghai, China, in 2019. He was a Postdoctoral Fellow associated with Dr. Amir Khajepour at the Department of Mechanical and Mechatronics Engineering, University of Waterloo, Waterloo, ON, Canada, from Jan.2020 to March.2021. He is currently an Assistant Project Scientist with the Department of Civil and Environmental Engineering, University of California, Los Angeles, CA, USA. His research interest includes state estimation, cooperative localization, cooperative perception, and dynamics control of the autonomous vehicle.
\end{IEEEbiography} 

\begin{IEEEbiography}[{\includegraphics[width=1in,height=1.25in,clip,keepaspectratio]{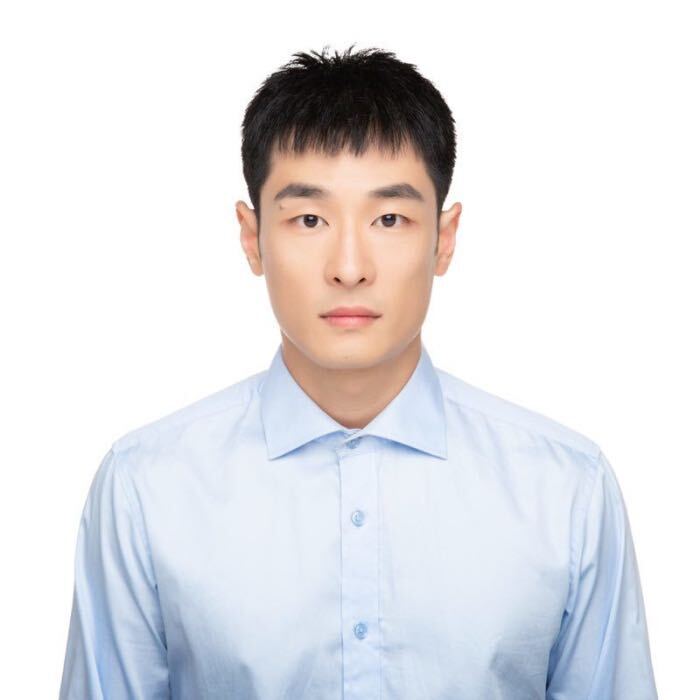}}] {Xu Han} received his B.S. in Electrical Engineering from the University of Cincinnati and his M.S. degree in Electrical Engineering from Washington University in St.Louis. He is currently a Ph.D. candidate at the University of California, Los Angeles, with the UCLA Mobility lab, advised by Prof. Jiaqi Ma. His latest research focuses lie in V2X cooperation, V2V cooperation, and high-level behavior planning for autonomous vehicles.
\end{IEEEbiography} 
\begin{IEEEbiography}[{\includegraphics[width=1in,height=1.25in,clip,keepaspectratio]{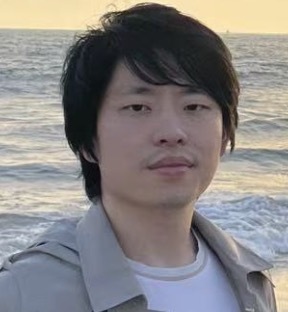}}] {Zonglin Meng} received his B.S. degree in Computer Science from University of Wisconsin Madison, Wisconsin, USA, in 2019. He is currently pursuing a Ph.D. degree with UCLA mobility Lab, University of California, Los Angelos. His research interests include computer vision, detection, tracking, and HD mapping.
\end{IEEEbiography} 

\begin{IEEEbiography}[{\includegraphics[width=1in,height=1.25in,clip,keepaspectratio]{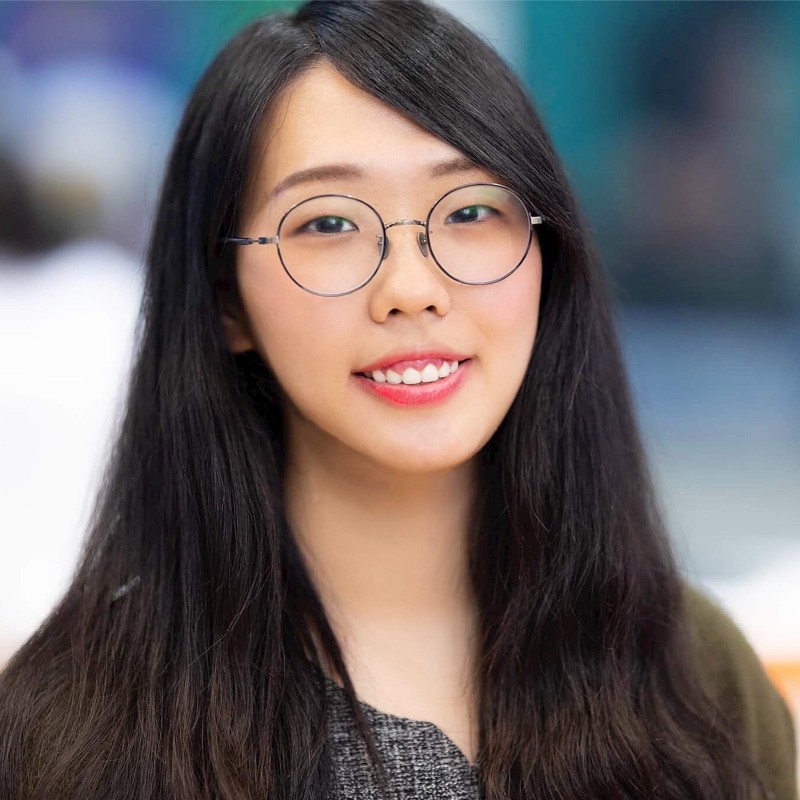}}] {Chia-Ju Chen} received the B.S. in Electrical and Computer Engineering from National Tsing Hua University, Taiwan at 2017 and M.S. from the University of Texas at Austin, USA at 2019. She is currently pursuing a Ph.D. degree at UCLA mobility lab. Her research interests include privacy, computer vision, and autonomous driving. 
\end{IEEEbiography}

\begin{IEEEbiography}[{\includegraphics[width=1in,height=1.25in,clip,keepaspectratio]{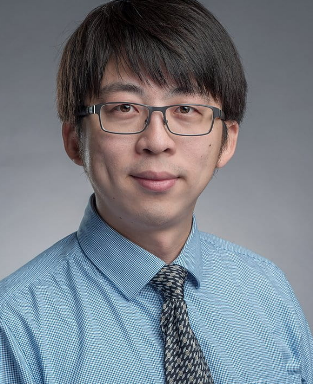}}]{Jiaqi Ma} received the Ph.D. degree in Transportation Engineering from University of Virginia, Virginia, USA, in 2014. He is currently an  Associate Professor at the UCLA Samueli School of Engineering and faculty lead of the New Mobility program at UCLA Institute of Transportation Studies. His research interests include intelligent transportation systems, autonomous driving, and cooperative driving automation. He is a member of the TRB Standing Committee on Vehicle-Highway Automation, the TRB Standing Committee on Artificial Intelligence and Advanced Computing Applications, and the American Society of Civil Engineers (ASCE) Connected and Autonomous Vehicles Impacts Committee, and the Co-Chair of the IEEE ITS Society Technical Committee on Smart Mobility and Transportation 5.0. He is Editor in Chief of the \textsc{IEEE Open Journal of Intelligent Transportation Systems}.
\end{IEEEbiography}

\end{document}